\documentclass[journal]{IEEEtran}
\usepackage{balance}
\usepackage{graphicx}
\usepackage{amsmath}
\usepackage{amsfonts,amssymb}
\usepackage{multirow}
\usepackage{makecell}
\usepackage{color}
\usepackage{array}
\usepackage{booktabs}
\usepackage{url}

\makeatletter
\newcommand{\rmnum}[1]{\romannumeral #1}
\newcommand{\Rmnum}[1]{\expandafter\@slowromancap\romannumeral #1@}
\makeatother

\ifCLASSINFOpdf
\else
\fi
\begin{document}

	\title{Tiny Obstacle Discovery by Occlusion-aware Multilayer Regression}

	\author{Feng~Xue,~
		Anlong~Ming~
		and~Yu~Zhou,~\IEEEmembership{Member,~IEEE} 
		\thanks{This work was supported by the National Natural Science Foundation of China 61703049 and the BUPT Excellent Ph.D. Students Foundation CX2020114. (Corresponding author: Yu Zhou)}
		\thanks{F. Xue and A. Ming are with \textcolor{black}{School of Computer Science}, Beijing University of Posts and Telecommunications, Beijing 100876, China
	    (e-mail:\{xuefeng,mal\}@bupt.edu.cn).}
		\thanks{Y. Zhou is with School of Electronic Information and Communications, Huazhong University of Science and Technology, Wuhan 430074, China (e-mail: yuzhou@hust.edu.cn)}}
	\maketitle

\begin{abstract}
Edges are the fundamental visual element for discovering tiny obstacles using a monocular camera.
Nevertheless, tiny obstacles often have weak and inconsistent edge cues due to various properties such as small size and similar appearance to the free space,
making it hard to capture them.
To this end, we propose an occlusion-based multilayer approach,
which specifies the scene prior as multilayer regions and utilizes these regions in each obstacle discovery module,
i.e., edge detection and proposal extraction.
Firstly,
an obstacle-aware occlusion edge is generated to accurately capture the obstacle contour by fusing the edge cues inside all the multilayer regions,
which intensifies the object characteristics of these obstacles.
Then, a multistride sliding window strategy is proposed for capturing proposals that enclose the tiny obstacles as completely as possible.
Moreover,
a novel obstacle-aware regression model is proposed for effectively discovering obstacles.
It is formed by a primary-secondary regressor, which can learn two dissimilarities between obstacles and other categories separately,
and eventually generate an obstacle-occupied probability map.
The experiments are conducted on two datasets to demonstrate the effectiveness of our approach under different scenarios.
And the results show that the proposed method can approximately improve accuracy by 19\% over FPHT and PHT,
and achieves comparable performance to MergeNet.
Furthermore, multiple experiments with different variants validate the contribution of our method.
The source code is available at \textit{\url{https://github.com/XuefengBUPT/TOD_OMR}}
\end{abstract}
	
\begin{IEEEkeywords}
Obstacle Discovery, Occlusion Edge, Object Proposal, Obstacle-Aware, Regression.
\end{IEEEkeywords}
	
\IEEEpeerreviewmaketitle
	
\section{Introduction}
\IEEEPARstart{A}{ccidents} caused when self-driving vehicles hit obstacles while driving cause personal and property damage.
To prevent such accidents,
it is necessary for autos to avoid running over various obstacles, such as bricks, lost cargo, and broken tires.
Extensive methods have been studied for obstacle discovery by utilizing different sensors, such as laser lidar,
time-of-flight sensors,
and cameras \cite{MergeNet,Pinggera2016Lost,TIP_ObsD,RBM}.
In recent years,
monocular cameras with a high spatial resolution have been popular with the rapid development of computer vision technology.
As a branch of object discovery \cite{WO,BING,TIP_adobe},
there are many methods for vision-based obstacle discovery, falling into three main categories:

\subsubsection{Geometric-based methods}
These methods exploit pixels in 3D space to find an obstacle surface.
Pinggera \emph{et al.} \cite{Pinggera2016Lost} applied statistical hypothesis testing to assess the drivable area and obstacle hypotheses.
Conrad \emph{et al.} \cite{ConradICRA} proposed a modified EM algorithm,
which classifies pixels to the ground or the obstacle by using the relative depth in different views.
Zhou \emph{et al.} \cite{ZhouICRA} \cite{ZhouICIP} proposed a search method for the dominant homography of the ground.
However,
the low accuracy of geometric cues limits these methods in discovering tiny obstacles in a long range.
	
\subsubsection{Segmentation-based methods}
These methods segment the pixels of obstacles by utilizing a deep neural network,
which has seen heavy use in recent years.
Ramos \emph{et al.} \cite{UON} designed an unexpected obstacle network based on fully convolutional networks \cite{FCN},
and combined the network output with geometric-based predictions \cite{Pinggera2016Lost}.
Gupta \emph{et al.} \cite{MergeNet} proposed a one-stage segmentation network trained by only 135 training images and achieved an excellent performance.
Although these pixelwise segmentation networks outperformed others by a large margin,
they require high-performance hardware.
Another issue is that the scarce features of tiny obstacles limit the accuracy of these methods.
	
\subsubsection{Proposal-based methods}
These methods first generate many bounding boxes for objects,
and then rerank these boxes.
Prabhakar \emph{et al.} \cite{TEN} generated a large number of proposals by Faster R-CNN \cite{FasterRCNN},
and applied a support vector machine to detect obstacles.
Garnett \emph{et al.} \cite{Garnett2017} combined the depth cues given by a 2D lidar and the object proposals acquired by SSD \cite{SSD} to locate the obstacles in 3D space.
Mancini \emph{et al.} \cite{JMOD2} constructed a joint network to simultaneously regress obstacle proposals and estimate scene depth.
However, due to the challenges in feature extraction and proposal generation,
these methods unable to address the problem of tiny obstacle discovery.
Our method follows a similar structure.
	
In addition to the above methods with clear categories,
several methods discover tiny obstacles in other ways.
Kristan \emph{et al.} \cite{cyber_FIOD} imposed weak structural constraints in marine scenarios,
and extracted an obstacle mask by a Markov random field framework.
Kumar \emph{et al.} \cite{MRF} proposed several features related to indoor scenes and segmented small obstacles with several assumptions.
Although these methods are not tailormade for road scenes,
they provide some inspiration,
namely, the prior information of a specific scenario deserves attention in obstacle discovery.

Considering the balance between hardware requirements and performance,
this paper focuses on a proposal-based method.
As described above, it is generally acknowledged that tiny obstacle discovery is challenging for the following two intrinsic reasons:
\begin{enumerate}
\item Due to the small size of obstacles and similar color to the road plane,
it is difficult to extract sufficient features of these obstacles.
\item When using proposal-based methods to capture obstacles,
there is always a blank space between two neighboring sliding windows,
which misses the proposals enclosing tiny obstacles tightly.
In turn,
an excessive intensive proposal sampling strategy leads to an increase in computational complexity.
\end{enumerate}
	
\begin{figure}
		\centering
		\includegraphics[width=1\linewidth]{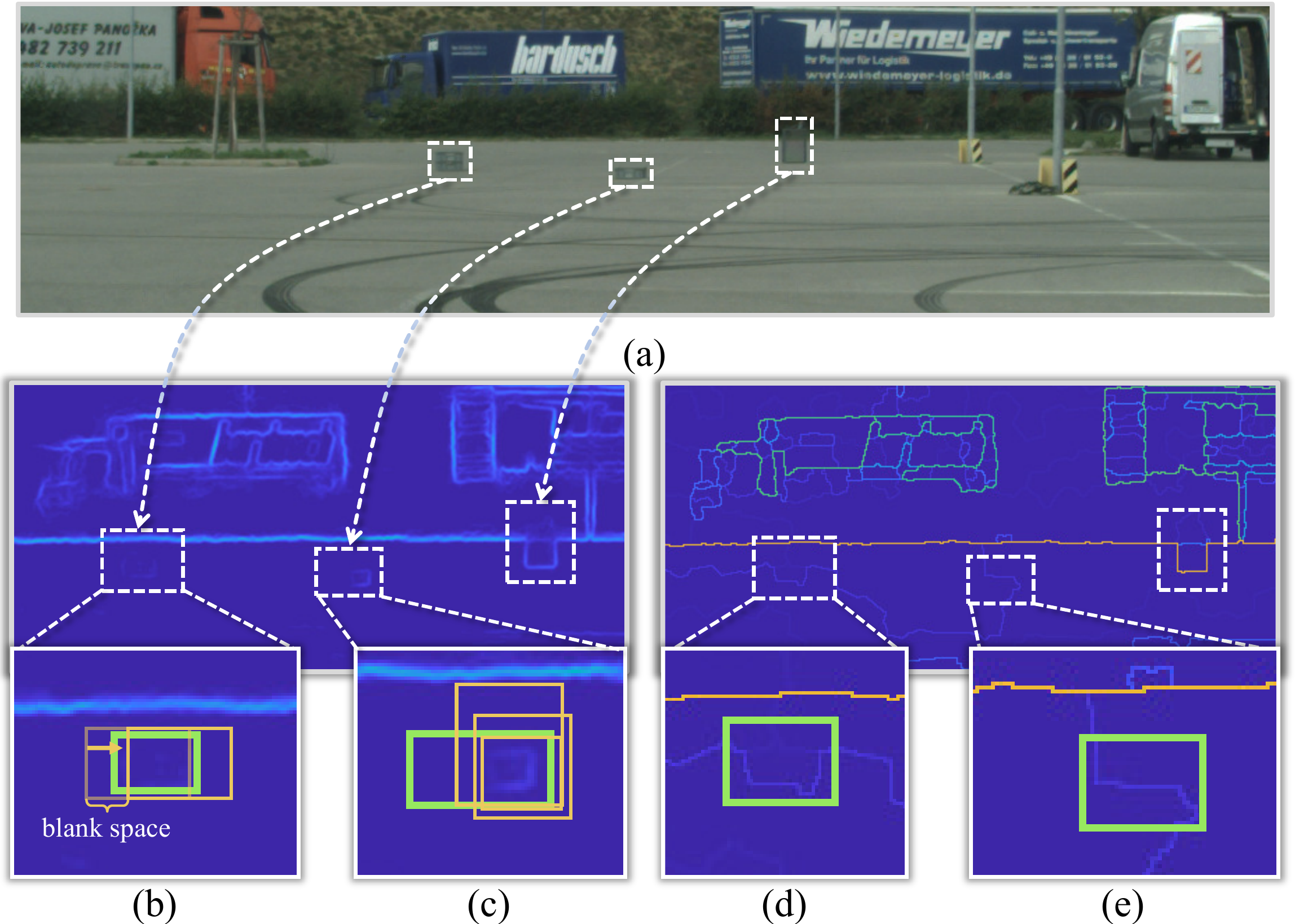}
		\caption{
			(a) shows an RGB image of a road scene.
			(b) shows the sliding window strategy in \textit{edge boxes}.
			(c) shows the inconsistent edge cues;
			the yellow boxes have a higher score than the green box.
			(d) and (e) show the occlusion edge generated by \cite{IS}.
			The four maps (b)(c)(d)(e) indicate the enlarged view of the maps above.
			The white boxes locate the obstacles.
			}
	\label{fig:rw1}
\end{figure}

In recent years,
several works \cite{OLP,TIP_SAPOPN,TRID} have attempted to solve the first issue from various aspects.
Among them,
Ma \emph{et al.} \cite{OLP} imposed an occlusion edge to capture object contours.
Compared with other edge cues,
due to the 3D cue of objects revealed by the occlusion edge,
the contours of small objects are captured with a higher response, thus boosting the edge-based proposal.
Thus, the occlusion edge is also helpful for tiny obstacle discovery.
However,
the first challenge mentioned above is still unresolved.
In some cases, such as Fig. \ref{fig:rw1}(b) and Fig. \ref{fig:rw1}(c),
the edge cues of obstacles in the long range are weak and inconsistent.
Hence, the occlusion edge is insufficiently extracted,
so that the occlusion-based proposal methods, such as \textit{object-level proposal} \cite{OLP}, cannot discover these tiny obstacles.

In this paper,
to address this remaining problem,
the scene prior (namely, the 2D spatial distribution of obstacles) is specified as several multilayer (ML) regions.
Based on these ML regions,
we enhance the edge cues of tiny obstacles and further generate the obstacle-aware occlusion edge that more completely fits the obstacle contour.
To alleviate the second issue mentioned above,
a multistride sliding window scheme is proposed, i.e.,
a dense sampling proposal in the high-layer region,
and a sparse sampling proposal in the low-layer region.
Furthermore,
to achieve a high discovery performance,
we explore two appearance properties of obstacles:
(1) the dissimilarity between the obstacle and road plane,
(2) the dissimilarity between the obstacles and other objects.
Then, an obstacle-aware regression model is designed to learn these dissimilarities and generate an obstacle-occupied probability map,
which consists of a pair of primary and secondary regressors.
The main contributions of this paper can be summarized as follows:
\begin{enumerate}
	
\item To eliminate occlusion edge detection failure caused by weak and inconsistent edge cues,
an obstacle-aware occlusion edge is designed by utilizing several multilayer regions.
\item A multistride sliding window strategy is proposed to
alleviate the failure in capturing tiny obstacles, 
and overly dense sampling is avoided in the full image.
\item To sufficiently learn two dissimilarities between obstacles and other elements,
a primary-secondary regressor is designed to generate an obstacle-occupied probability map.
The proposed method achieves state-of-the-art performance on two datasets \cite{Pinggera2016Lost}\cite{BOVCON2018}.

\end{enumerate}
	
\section{Related work}
\label{sec:2}
An earlier version of this work appeared in \cite{ICRA}.
Since our method closely depends on \textit{edge boxes} \cite{EB} and \textit{object-level proposals} \cite{OLP},
they are briefly introduced below.

\subsection{Reviewing Edge boxes}
To model the observation of objects in an image,
\textit{edge boxes} \cite{EB} densely search bounding boxes in the image by the sliding window strategy and define a specific objectness score based on the edge probability map of this image.
	
However,
due to the low resolution of tiny obstacles,
these obstacles occupy few pixels and have a similar color distribution with the road,
which leads to blurry and weak obstacle edges,
as shown in Fig. \ref{fig:rw1}(b) and Fig. \ref{fig:rw1}(c).
In \textit{edge boxes} \cite{EB},
the box intersecting the weak edge obtains a higher score ranking,
as in the yellow boxes shown in Fig. \ref{fig:rw1}(c).
Additionally, since there is no spatial constraint between pixel values in the edge probability map,
different edge pixels of the same obstacle have completely different probabilities,
making these obstacles less like objects.
Furthermore, inferring from the \textit{objectness} \cite{WO},
these two inherent reasons give rise to low objectness of the tiny obstacle proposals.
Hence, it is necessary to design a method to improve edges with a closed region.
However,
as shown in Fig. \ref{fig:rw1}(b),
due to the small size of tiny obstacles,
the sliding window easily ignores the bounding box that tightly encloses the obstacle.

\begin{figure*}
		\centering
		\includegraphics[width=0.91\linewidth]{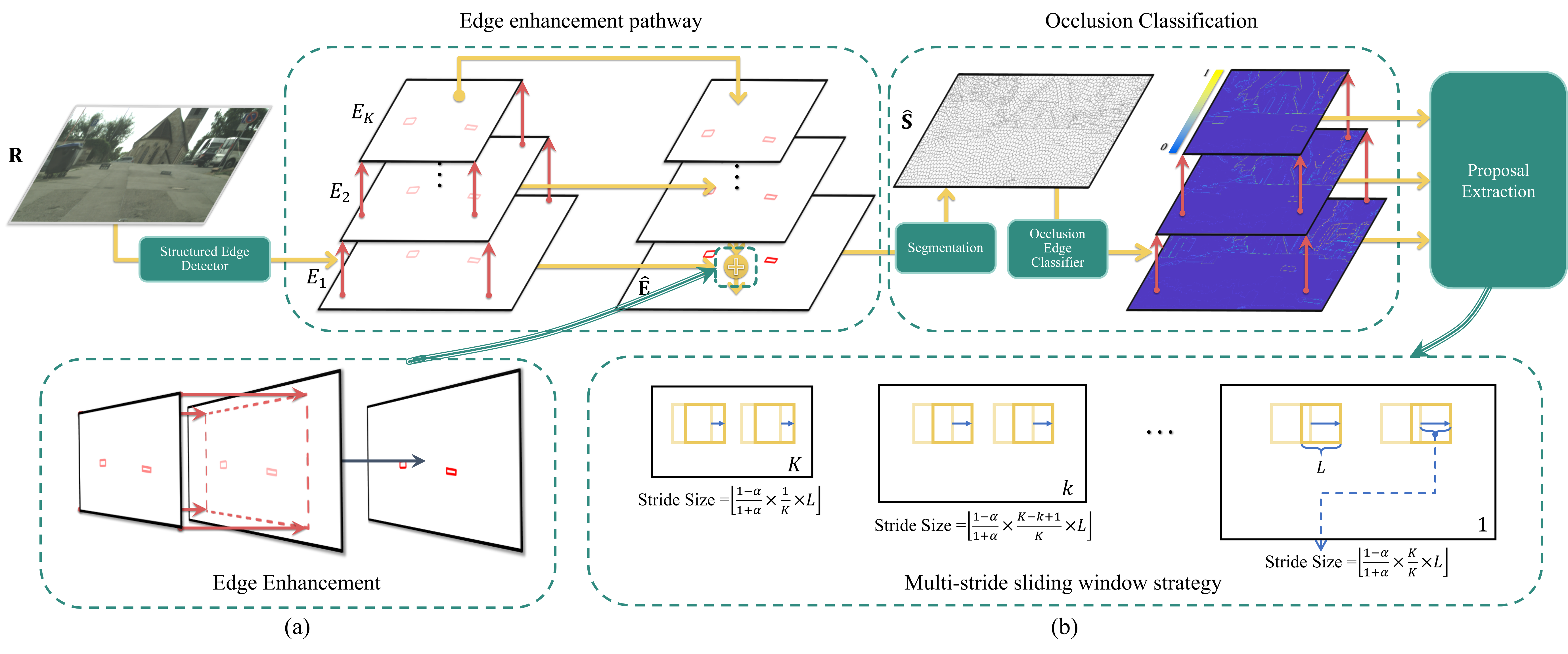}
		\caption{
The pipeline of our proposed approach
(a) shows the edge enhancement detail.
(b) indicates the multistride sliding window strategy.}
		\label{fig:Meth1}
\end{figure*}

\subsection{Reviewing Object-level Proposal}
To discover the low-ranking proposal of obstacles in \textit{edge boxes} \cite{EB},
the \textit{object-level proposal} \cite{OLP} constructs an occlusion-based objectness that considers the surface cue.
The occlusion edge \cite{IS} presents the edge revealing the depth discontinuity between the object and background.
It takes the edges between adjacent regions of an oversegmented image as candidates and classifies the candidate edges into two subsets:
occlusion edges and trivial edges.
Compared to the typical edge cues \cite{SED2,UCM,HED},
the occlusion edge has a more robust response to the object contour,
especially for small objects.
As shown in the map in Fig. \ref{fig:rw1}(d) and Fig. \ref{fig:rw1}(e),
by considering the surface cue,
the rightmost obstacle acquires a high response.
Hence, the \textit{object-level proposal} is more appropriate for tiny obstacle discovery than others.
	
However, the \textit{object-level proposal} may not fare well in practice.
Some tiny obstacles in the long range rarely acquire a sufficient occlusion edge,
leading to detection failure,
as shown in the second row of Fig. \ref{fig:rw1}(c).
Intrinsically,
since these obstacles vary in scale and have similar appearances with the road plane,
they have weak and inconsistent edges,
making it hard to apply the candidate edge with obstacle contours.
In other words,
the contour of the obstacle is not considered as candidate edges and thus fails to be detected.
In this case,
the obstacle cannot be represented as an object,
which brings great difficulty in tiny obstacle discovery.
Our method is designed to address this challenge.

\subsection{The Differences with the Earlier Version}
The main differences between this paper and the earlier version \cite{ICRA} are summarized as follows:
\begin{enumerate}

\item To overcome the difficulty in capturing tiny obstacles by the existing sliding window,
we propose a novel multistride sliding window strategy to capture tiny obstacles as completely as possible.
\item Different from the regressor in \cite{ICRA},
a primary-secondary regressor is designed to learn two types of dissimilarity for addressing the insufficient learning of a single model.

\item This paper proposes a training sample
selection scheme to improve the performance of the regression model comprehensively.
\item This paper slightly adjusts the scheme in edge enhancement to reduce the redundant calculations.
\end{enumerate}

\section{Method}

The proposed method is organized as follows:
Section \ref{sec:MethodA} demonstrates the multilayer region extraction.
Section \ref{sec:MethodB} demonstrates how to extract obstacle-aware occlusion edges inside these extracted regions.
Section \ref{sec:MethodC} introduces the multistride sliding window strategy in the multilayer regions,
which improves the recall of the obstacle proposals.
Section \ref{sec:MethodD} introduces the obstacle-aware regression model,
which generates the obstacle-occupied probability map for obstacle avoidance.

\subsection{Offline Multilayer Region Extraction}
\label{sec:MethodA}

We first introduce the multilayer (ML) regions used in the following parts of our method:
occlusion edge detection and proposal extraction.
Different from the concept of multiresolution in the pyramid model,
these regions are expected to contain obstacles at different distances.
Therefore,
we utilize the spatial distribution of obstacles in the training set to determine ML regions.

\begin{figure}
		\centering
		\includegraphics[width=1\linewidth]{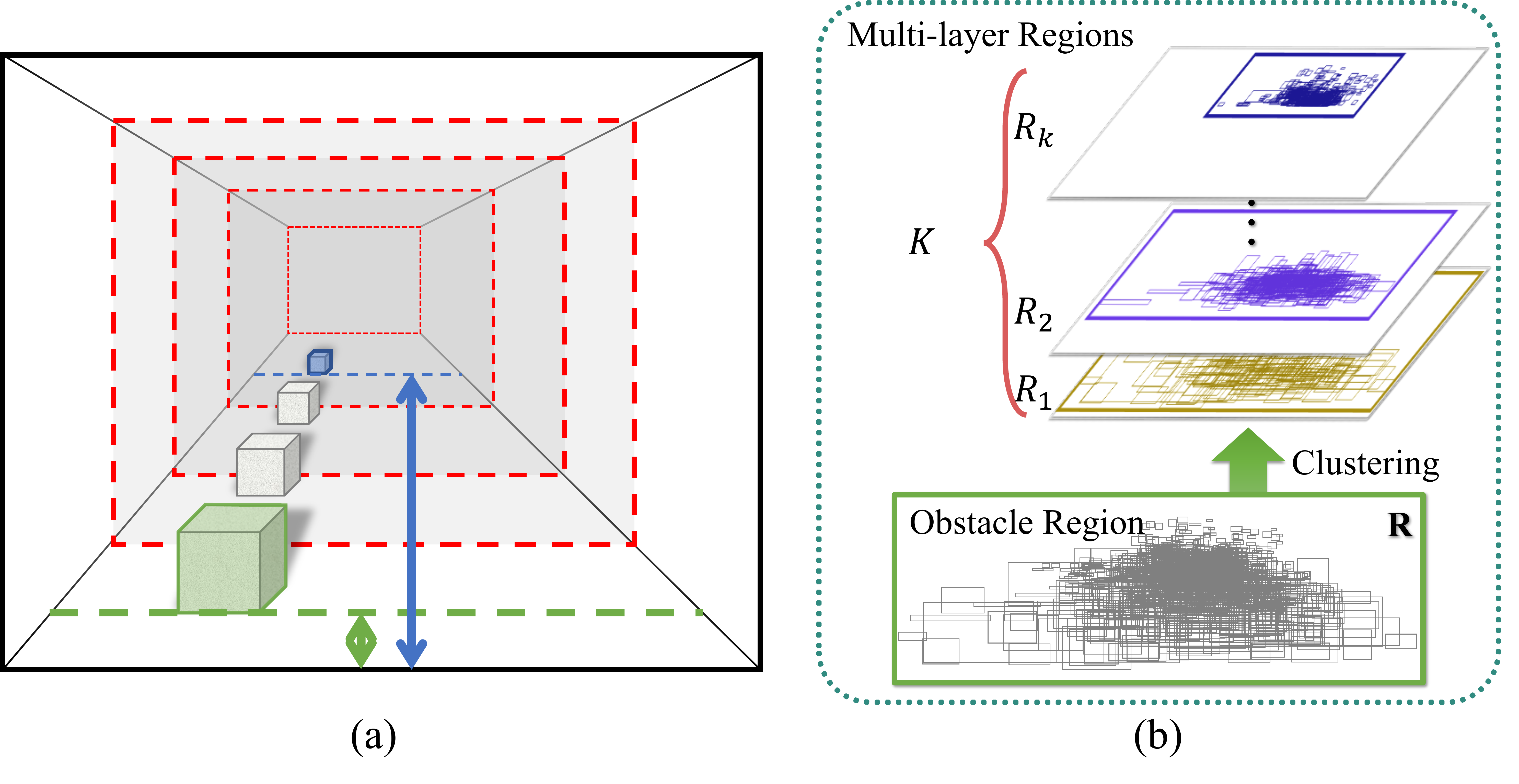}
		\caption{(a) A sketch of the perspective.
			For a fixed size object, the farther it is in 3D space,
			the smaller it is in the 2D image and farther away from the image bottom.
			(b) The green box $\textbf{R}$ encloses all the obstacles,
			and the rectangular boxes with different colors denote the ML regions.}
		\label{fig:Meth2}
\end{figure}

First,
a pseudodistance is designed to indicate the distance from an obstacle to the camera.
As the principle of perspective \cite{MV} shown in Fig. \ref{fig:Meth2}(a),
for a mounted camera and a fixed size obstacle in 3D space,
obstacles that are farther away are visually smaller and farther away from the bottom of the image.
Thus, pseudodistance is defined as a group of two properties:
(i) the pixel length from the obstacle bottom to the image bottom,
(ii) the number of pixels occupied by an obstacle.
Second,
given all training obstacles in the dataset, i.e., $O = \{o_1,o_2,...,o_N\}$,
the region enclosing all obstacles is denoted by the green rectangular box $\textbf{R}$.
Indicating by the pseudodistance,
the \textit{k-means clustering} groups all obstacles $O$ into the $K$ subset,
i.e., $\textbf{O} = \{O_1,O_2,...,O_K\}$,
as shown in Fig. \ref{fig:Meth2}(b).
Note that the obstacles in the same subset $O_k\in\textbf{O}$ have similar vertical locations and similar areas,
Most of the obstacles in $O_k$ are farther than those in $O_{k-1}$.
Finally, the ML regions are obtained by considering the partition of $\textbf{O}$:
\begin{itemize}
    \item The first ML region $R_1$ tightly encloses all the obstacles in $\{\textbf{O}\}$.
    \item The second ML region $R_2$ corresponds to $\{\textbf{O}\setminus O_1\}$.
    \item In the same way, the ML region $R_K$ contains the obstacles in $\{\textbf{O}\setminus O_1,O_2...,O_{K-1}\}$.
\end{itemize}
Generally, the farthest obstacle exists in the ML region $R_K$.

\subsection{Obstacle-aware Occlusion Edge}
\label{sec:MethodB}
Given an RGB image of a scene,
we first employ the \textit{structured edge detector} \cite{SED2} to obtain the edge probability inside the ML region $R_1$,
as the whole pipeline illustrated in Fig. \ref{fig:Meth1}.
Then, the edge cue of tiny obstacles is enhanced by the edge enhancement pathway.
Eventually, the occlusion edge is extracted by utilizing the enhanced edge.

\subsubsection{Edge Enhancement Pathway}

Let $E_1$ denote the edge probability inside the ML region $R_1$.
Along the ML regions $R_1,R_2,...,R_K$,
the whole edge map $E_1$ is clipped into the corresponding edge probability maps $\{E_1,E_2,...,E_K\}$.
Since $R_{k}\cap R_{k+1}=R_{k+1}$,
and the region $R_{k+1}$ indicates a farther scene than $R_{k}$,
all edge maps $E_k$ are added together,
which increases the edge probability of overlapping areas.
In detail,
all pixels in map $E_k$ are summed to the corresponding pixels in map $E_{k-1}$,
enhancing the edge cues in the overlap hierarchically.
In this way,
the pyramid bottom, i.e., the enhanced edge probability map $\hat{\bf E}$, is generated.
Unlike \cite{ICRA},
only the bottom map is reserved in this paper,
because this map $\hat{\bf E}$ expresses the obstacle contour more clearly than others.
The $1$-st rows of Fig. \ref{fig:Meth4}(b) and Fig. \ref{fig:Meth4}(c) , respectively, depict the edge probabilities obtained by the previous method and our method.
Observably,
in the former,
the plastic case's edge probability is close to zero,
therefore, the case is hardly discovered.
In the latter,
our method increases the edge of this case several times,
making it much higher than zero and more accessible for discovery.

\subsubsection{Occlusion Edge Detection}
\begin{figure}
		\centering
		\includegraphics[width=1\linewidth]{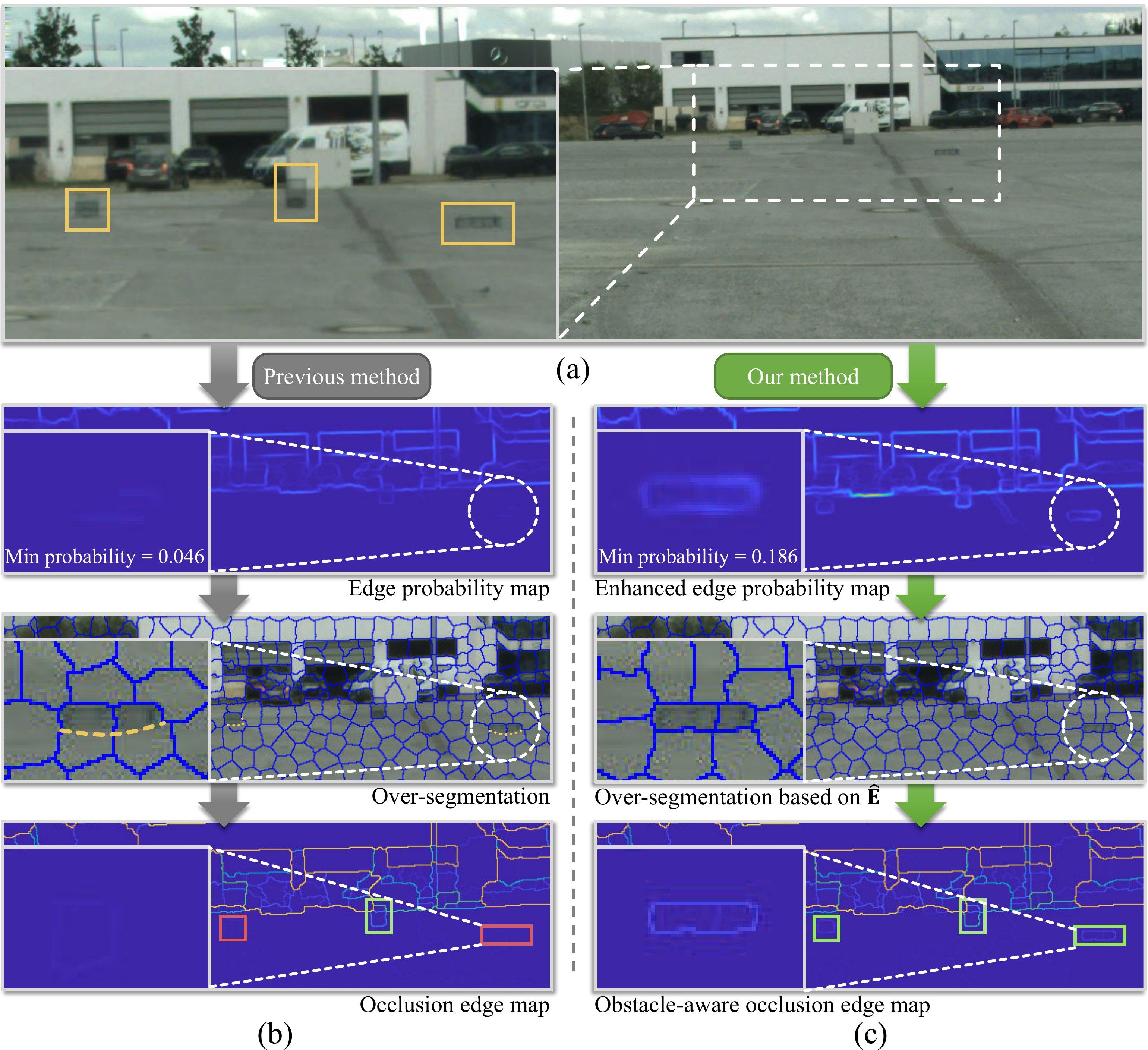}
		\caption{
			(a) represents an RGB image of a road scene, and the yellow boxes show the obstacles.
			(b) represents the edge cues, segmentation, and occlusion edge generated by the previous works \cite{IS,ICIP15}.
            The yellow dotted lines indicate the lost contours,
			and the blue lines indicate the boundaries of all atomic regions.
			(c) represents the results of our method.
			The red boxes enclose the false-negative contours,
			and green correspond to true positive ones.
		}
		\label{fig:Meth4}
\end{figure}

To represent obstacles by occlusion edge,
the scene is first oversegmented into many atom regions by \cite{SP},
where the enhanced edge probability map $\hat{\bf E}$ is used as a dominant element.
Subsequently, the \textit{occlusion classifier} \cite{IS} is employed to find the occlusion edge from the boundaries of these atom regions.
Naturally,
due to the dependence of the occlusion edge on oversegmentation,
once the weak edge of the tiny obstacle cannot be fitted by these atom regions,
it cannot be detected by the occlusion classifier.
The $2$-nd row of Fig. \ref{fig:Meth4}(b) represents the atom regions of the previous method,
and the $3$-rd row corresponds to the occlusion edge map.
It is clear that the edge of the plastic case is not fitted by the atom region obtained in the previous work,
and thus is undetected.
In contrast,
as shown in the $2$-nd and $3$-rd rows of Fig.\ref{fig:Meth4}(c),
the enhanced map of our method improves the oversegmentation.
Thus, the contour of this case is detected.

\begin{figure*}
	\centering
	\includegraphics[width=0.95\linewidth]{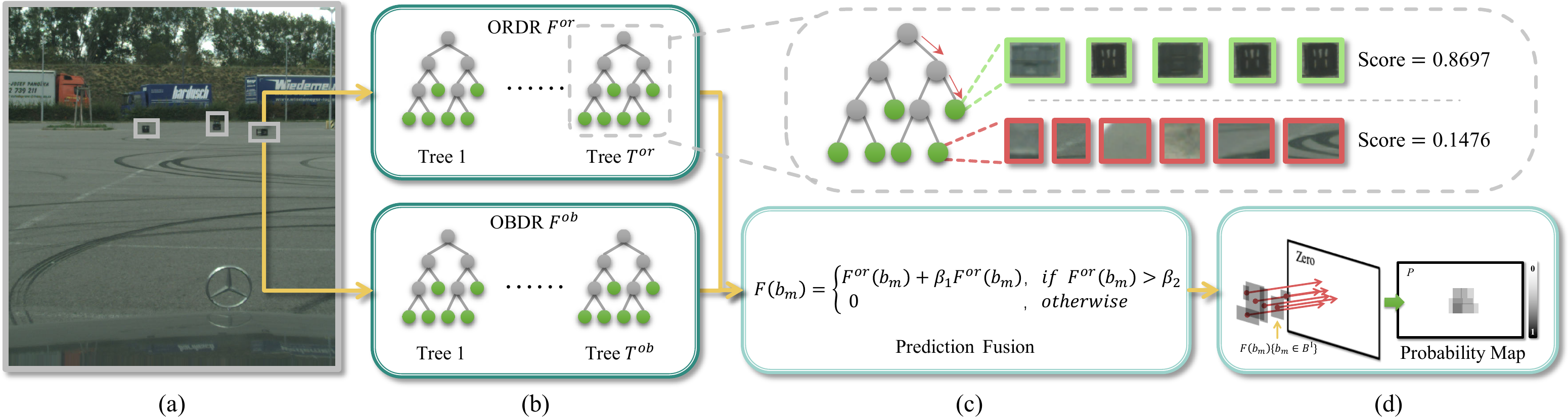}
	\caption{
		The pipeline of the obstacle-aware regression model is represented above.
		(a) represents an RGB image of the road scene, and the proposals enclosing obstacles
		are marked in gray.
		(b) represents the two regressors,
		and the gray dotted rectangle shows the enlarged view of a tree.
		(c) represents the prediction fusion scheme.
		(d) represents the generation of the obstacle-occupied probability map,
		where a semitransparent black block is a proposal.
	}
	\label{fig:Meth5}
\end{figure*}

\subsection{Multistride Sliding Window Strategy}
\label{sec:MethodC}
With the representation in \cite{EB},
$\alpha$ denotes a fixed intersection over union (IoU) between two neighboring boxes,
determining the stride of a sliding window.
However,
although we completely capture the contours of tiny obstacles,
it is still hard to capture obstacles with a sliding window,
because the stride with a fixed IoU makes most of the boxes intersect with the tiny obstacle, rather than tightly enclose it.

To address this issue,
we design a multistride sliding window strategy,
i.e., more densely taking boxes in a low-layer region and sparsely taking boxes in a high-layer region.
Specifically,
the sliding window size ranges from 100 pixels to the full ML region,
and the aspect ratio ranges from $1/3$ to $3$.
As shown in Fig. \ref{fig:Meth1}(b),
the stride of a sliding window translating in the X or Y direction is expressed as $\lfloor\frac{1-\alpha}{1+\alpha}\times\frac{K-k+1}{K}\times L\rfloor$,
where $K$ is the whole number of ML regions,
$k$ is the region number selected for sampling,
$L\in\{h_W,w_W\}$ denotes the height or width of the search window,
and $\frac{1-\alpha}{1+\alpha}\times\frac{K-k+1}{K}$ denotes the stride ratio.
The smallest stride is equal to 2 pixels.
Compared with the previous sampling strategy \cite{EB},
we reduce the stride in the $k$-th ML region $R_k$ by $\frac{K-k}{K}$ times.
Thus, this strategy applies proper strides for obstacles of different sizes,
which better captures the proposals of tiny obstacles.

By utilizing the sampling strategy,
the occlusion edge maps of all the layers are employed to generate the proposals.
Consequently,
we consider the proposals in all ML regions as an initial set, i.e., $B^{\bf I}$.
In addition, all the proposals in set $B^{\bf I}$ are predefined to three categories:
(\rmnum{1}) road area, (\rmnum{2}) obstacle, (\rmnum{3}) non-road background.
For the road proposals,
more than 50\% of the pixels belong to the road,
and other proposals are similar.

\subsection{Obstacle-aware Regression Model}
\label{sec:MethodD}
Given the initial set of proposals $B^{\bf I}$,
this section first characterizes these proposals.
Then, a regression model is learned to find the proposals that may contain an obstacle.
Eventually, an obstacle-occupied probability map is generated to avoid obstacles.

\subsubsection{Proposal Expression}
Several features represent the proposal:
(\rmnum{1}) \textit{Edge and structure}:
Edge Density (ED) \cite{WO},
average, maximum, and mode of edge response,
the ratio of the mode measures the statistical information of the edge;
ED measures the density of edges near the box borders.
(\rmnum{2}) \textit{Pseudodistance}:
Following \cite{MCG}, size, position, height, width, and aspect ratio of the proposal;
The combination of these features is associated with pseudodistance.
(\rmnum{3}) \textit{Objectness score}:
Following \cite{WO},
the occlusion-based objectness score measures the likelihood that a box contains an object.
(\rmnum{4}) \textit{Color}:
Color contrast (CC) \cite{WO} and color variance (CV) of the proposal;
CC measures the color dissimilarity of a box to its immediate surrounding area,
and the CV of a box in the HSV image reflects the color dispersion inside this box.
The cosine distance between the HSV histograms is employed as the metric of CC.
As shown in Table \ref{table:features},
all the features can be easily calculated.

For the proposal $b^{\bf I}_m \in B^{\bf I}$,
stacking all the features,
a 20-dimensional feature vector $v_{b^{\bf I}_m}\in \mathbb{R}^{20}$ (7 for edge and structure, 6 for pseudodistance, 1 for objectness score, 6 for HSV color space) is constructed.

\begin{table}[!tp]
	\begin{center}
		\caption{
			The features of proposal in this paper.
		}
		\begin{tabular}{c c c}
			\Xhline{1.2pt}
			{\bf Category} & {\bf Feature name} & {\bf Count} \\
			\hline
			\multirow{7}{*}{{\it Edge and structure}} & max edge response & \multirow{7}{*}{7} \\
			& most edge response & \\
			& proportion of most response & \\
			& average edge response & \\
			& average edge response in the inner ring & \\
			& edge density & \\
			& edge density in the inner ring  & \\
			\hline
			\multirow{6}{*}{{\it Pseudodistance}} & normalized area & \multirow{6}{*}{6} \\
			& aspect ratio & \\
			& X coordinate of the center & \\
			& Y coordinate of the center & \\
			& width & \\
			& height & \\
			\hline
			{\it Objectness Score} & occlusion-based objectness & 1 \\
			\hline
			\multirow{6}{*}{{\it Color}} &  color variance in the H channel & \multirow{6}{*}{6} \\
			& color variance in the S channel & \\
			& color variance in the V channel & \\
			& color contrast of the H channel & \\
			& color contrast of the S channel & \\
			& color contrast of the V channel & \\
			\Xhline{1.2pt}
		\end{tabular}
		\label{table:features}
	\end{center}
\end{table}	

\subsubsection{Model Structure}

Based on the \textit{random forest} \cite{DF},
an obstacle-aware regression model consists of two regressors,
i.e., the obstacle-road dissimilarity regressor (ORDR) with a primary role and the obstacle-background dissimilarity regressor (OBDR) with a secondary role.
The core insight behind the regression model is that learning the appearance dissimilarity between obstacles and monotonous roads,
combined with the dissimilarity learned from obstacles and other objects can jointly improve obstacle discovery performance.

The two regressors are denoted as $F^{or} = \{f^{or}_i|i=1,\dots,T^{or}\}$ and $F^{ob} = \{f^{ob}_i|i=1,\dots,T^{ob}\}$.
$f^{or}_i$ and $f^{ob}_i$ denote the trees in ORDR and OBDR,
and they follow the typical binary tree structures.

As shown in Fig. \ref{fig:Meth5},
each tree consists of internal nodes and leaf nodes.
The internal node classifies the proposals reaching this node,
and passes these proposals to its left or right child node until reaching a leaf node.
Moreover, the reached leaf node stores a score that is provided to the input proposal.
As shown in Fig. \ref{fig:Meth5},
in the forest,
the training proposals that fall in the same leaf node have a similar appearance.
It is observable that the proposals of distant cargos with square shapes reach the same leaf node.
For training the two regressors,
two elements,
i.e., training sample selection and feature selection,
are discussed below.

\subsubsection{Training Sample Selection}
\begin{figure}
\centering
\includegraphics[width=0.9\linewidth]{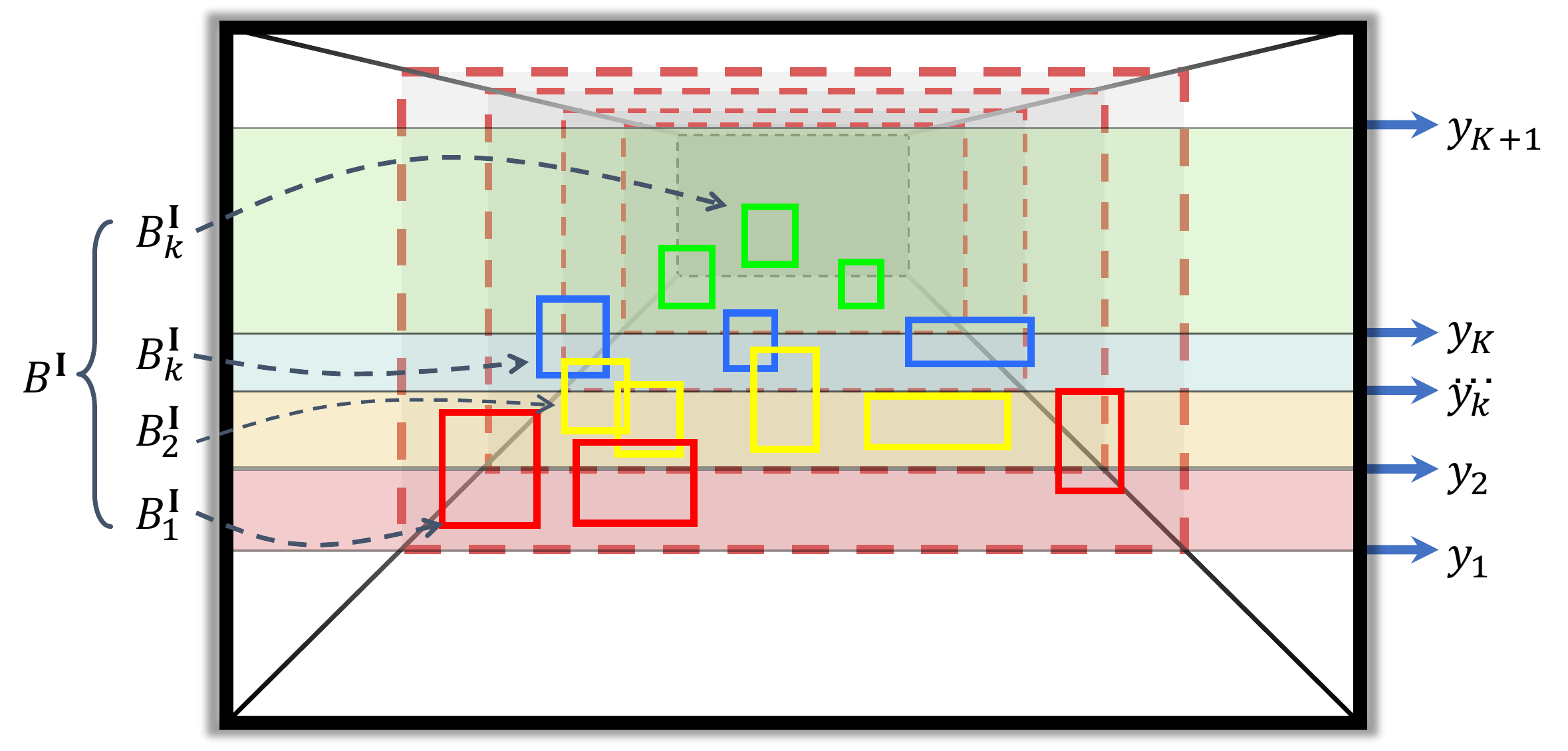}
\caption{
The sketch shows the vertical ranges of the generated proposals,
which are utilized in training sample selection.
}
\label{fig:Meth3}
\end{figure}

To sufficiently learn the obstacles,
a training sample selection strategy is introduced,
which guarantees the diversity of samples in terms of scale and objectness.
First, 
the whole scene is partitioned into several vertical ranges,
and training samples are taken from these ranges in proportion.
Specifically,
as depicted in Fig. \ref{fig:Meth3},
$[y_k,y_{k+1}]$ denotes the vertical range,
where $y_k$ is the vertical position of the $R_k$ bottom,
and $y_{K+1}$ is the top of $R_K$.
Then, let $B^{\bf I}_k$ denote a set of proposals whose bottom falls inside the vertical range $[y_k,y_{k+1}]$.
For the ORDR,
we randomly take ${\bf n}^-_k$ samples with an obstacle IoU less than 0.5 from set $B^{\bf I}_k$ as the training sample.
To guarantee the diversity of training samples in objectness
among the ${\bf n}^-_k$ samples,
$\tau_l$ is the ratio of samples with objectness larger than $\tau_o$.
For positive samples,
we take ${\bf n}^+_k$ training samples that have the top IoU with obstacles from set $\{B^{\bf I}_k\}$.

Note that the two regressors are trained by using different samples.
Based on the training sample selection scheme mentioned above,
the proposals $B^{or} = \{b^{or}_1,b^{or}_2,\dots,b^{or}_M\}, B^{or}\subset B^{\bf I}$,
containing (\rmnum{1}) and (\rmnum{2}),
are considered the training samples of ORDR.
For OBDR,
a similar scheme is employed to extract training samples $B^{ob} = \{b^{ob}_1,b^{ob}_2,\dots,b^{ob}_M\}, B^{ob}\subset B^{\bf I}$, containing (\rmnum{2}) and (\rmnum{3}).
Both regressors are designed to regress the overlap between each proposal and the ground truth.

\subsubsection{Prediction Fusion}
	For a proposal $b_m$,
	the prediction of each regressor is formulated as the average of each tree output:
\begin{equation}
\begin{split}
	F^{or}(b_m) = \frac{1}{T}\sum\nolimits_{i=1}^{T}f^{or}_i(v_{b_m}) \\
	F^{ob}(b_m) = \frac{1}{T}\sum\nolimits_{i=1}^{T}f^{ob}_i(v_{b_m})
\end{split}
\end{equation}
	where $f^{or}_i(v_{b_m})$ denotes the output of each tree in ORDR to proposal $b_m$,
	and $f^{ob}_i(v_{b_m})$ corresponds to OBDR.
	Since the OBDR only learns the difference between the obstacles and other objects,
	this assistant regressor fails to separate the obstacle from the road plane.
	Thus, the final score of a proposal $b_m$ is formulated as:
\begin{equation}
F(b_m) = 
\begin{cases}
F^{or}(b_m)+\beta_1 F^{ob}(b_m), &\text{if $F^{or}(b_m) > \beta_2$}\\
0, &\text{otherwise}
\end{cases}
\end{equation}
where $\beta_1$ denotes the OBDR weight,
$\beta_2$ is determined by the proposal with the highest score of the $\tau_\beta$ ratio,
and $F(b_m)$ denotes the final score of proposal $b_m$.
The ORDR removes many road proposals,
which immensely decreases the false positive prediction.
Moreover, the OBDR further raises the confidence of the obstacle proposal.
Thus, the fused predictions better represent the obstacles than our base approach \cite{ICRA}.

\subsubsection{Obstacle-Occupied Probability Map}
The scores of the top 50\% proposals in $B^{\bf I}$ are accumulated in the corresponding pixels to produce a probability map $P$:
\begin{equation}
P(pixel(\hat{p})) =
\frac{1}{\mathcal{N}^P}\sum\nolimits_{b_m\in B^r}\sum\nolimits_{\hat{p} \in b_m} F(b_m)
\end{equation}
where $pixel(\hat{p}) = (u_{\hat{p}},v_{\hat{p}})$ denotes the coordinate of pixel $\hat{p}$.
$\frac{1}{\mathcal{N}^P}$ denotes the normalization term.
If $\hat{p}$ is inside $b_m$,
the $b_m$ score $F(b_m)$ is summed into $P(pixel(\hat{p}))$.
Due to the obstacle enclosed by multiple proposals,
the obstacle area in $P$ has a higher response than the road.
Finally,
the pixels with high probability belong to the obstacle.
Thus,
by setting a threshold in the obstacle-occupied probability map,
an obstacle mask is provided for obstacle avoidance.

\section{Experiment}
In this section,
we evaluate the proposed method on two challenging datasets:
the \textit{Lost and Found} dataset (LAFD) and \textit{Marine Obstacle Detection} dataset (MODDv2) \cite{BOVCON2018}.
The results on LAFD are first reported to compare with state-of-the-art obstacle discovery methods,
and the best variants are evaluated on MODDv2.
Furthermore, the ablation study represents the effectiveness of our method.

\begin{figure*}
\centering
\includegraphics[width=1\linewidth]{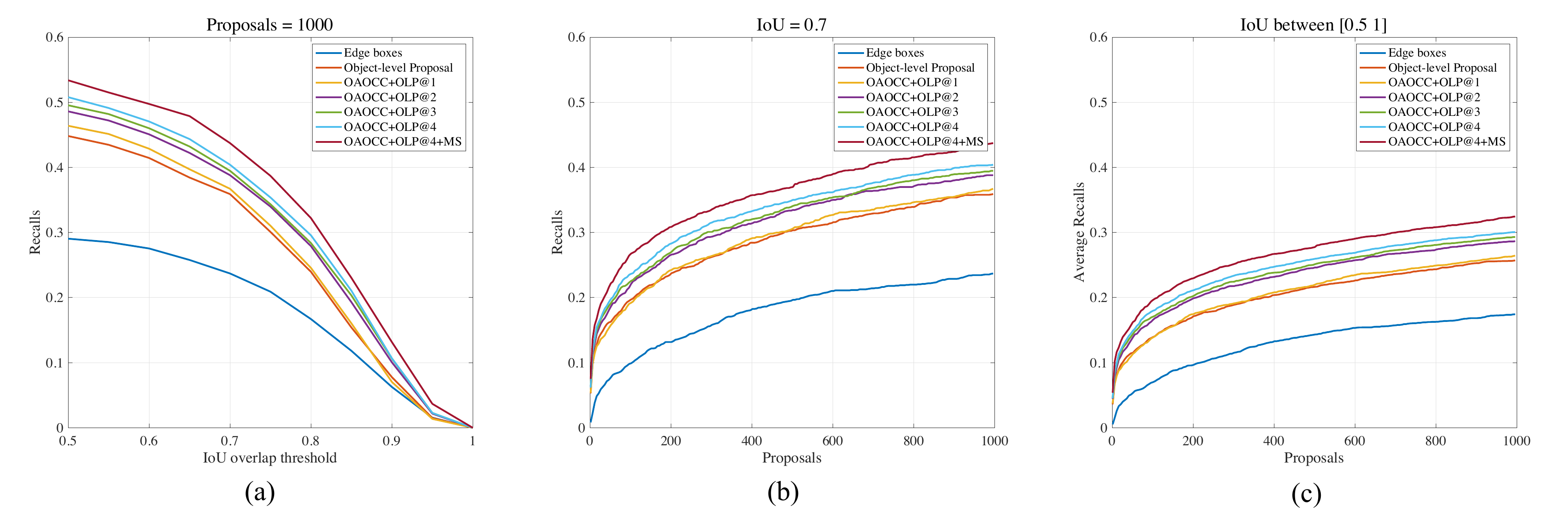}
\caption{
Instance-level recall performance of our variants with MS and OAOCC on the LAFD.
(a) Recall versus the number of proposals given the threshold of IoU is 0.7.
(b) Recall versus IoU threshold given 1,000 proposals.
(c) Average recall versus the number of proposals between IoU 0.5 to 1.}
\label{fig:recall1}
\end{figure*}

\subsection{Metrics}
\subsubsection{Pixel-level ROC}
Referring to \cite{Pinggera2016Lost},
the pixel-level receiver-operator-characteristic (ROC) curve compares the true positive rate (TPR) over the false positive rate (FPR), which is defined as follows:
\begin{equation}
\begin{split}
TPR = \frac{TP}{GT_{obstacle}}\\
FPR = \frac{FP}{GT_{road}}
\end{split}
\end{equation}
where $TP$ denotes the number of correctly discovered obstacle pixels and $FP$ denotes the number of road pixels that are incorrectly predicted as the obstacle pixels.
$GT_{obstacle}$ refers to the total pixels of the obstacle class, and $GT_{road}$ corresponds to the road area.
In this paper,
One hundred thresholds from 0 to 1 are averaged to segment the obstacle-occupied probability map,
the pixels over the threshold are labeled as obstacle pixels.

\subsubsection{Instance-level Recall}
For the instance-level metric,
the recall rate of the obstacle proposal is employed as a standard measure.
Our basic method \cite{ICRA} defines the IoU as the intersection over union (IoU) between the predicted proposals and the ground truth segmentation,
which focuses on a part of the obstacle.
However, it is improper for tiny obstacle discovery,
because we hope to discover a complete obstacle,
but not a part of it.
Thus,
this paper defines IoU as the intersection over union between predicted proposals and the ground truth bounding boxes,
which is more widely used in the works of object detection \cite{FasterRCNN,YOLO}, object proposal \cite{EB,OLP} and object discovery \cite{Kwak2015Unsupervised}.

Based on this definition,
three proposal metrics in \cite{OLP} are used to make comparisons on the recall rate for obstacle discovery.
\begin{itemize}
	\item Taking the top 1,000 proposals, the IoU threshold ranges from 0.5 to 1.
	\item Setting the IoU threshold to 0.7, the number of proposals ranges from 1 to 1,000.
	\item The average recall (AR) between IoU 0.5 and 1 is introduced, ranging the number of proposals from 10 to 1,000.
\end{itemize}

\subsection{Lost and Found Dataset}
\subsubsection{Dataset}
The \textit{Lost and Found} dataset (LAFD) is the only one publicly available dataset that focuses on discovering the small obstacles and lost cargo on roads.
The whole dataset records 13 different challenging street scenarios and 37 different obstacles,
and is split into a training subset and a testing subset,
containing 1,036 images in the training/validation set and 1,068 images in the testing set.
In the testing set,
there are some more complicated scenes and obstacles that do not appear in the training set to verify the compatibility of the methods.
The dataset provides images from a stereo camera,
and only the left camera is utilized to verify our method.

\subsubsection{Variants}
The obstacle-aware occlusion edge map (OAOCC) and multistride sampling strategy (MS) can be used to enhance the proposal.
OLP@k denotes extracting the object proposal from region 1 to region k.
Comparing the variants,
i.e., OAOCC+OLP@k+MS and OAOCC+OLP@k,
the experiment illustrates the contribution of each part to
the extraction of proposals.
Note that OAOCC+OLP@4 is used in our basic method \cite{ICRA}.
To assess the impact of the positive sample ratio on the performance of the ORDR,
three variants,
which have different positive sample ratios,
are defined as follows:
\begin{itemize}
\item $\{{\bf n}^+_1,{\bf n}^+_2,{\bf n}^+_3,{\bf n}^+_4\} = \{17,17,17,17\}$,
abbreviated as $\{17-17\}$
(regardless of which vertical range the obstacle is in,
it selects the top 17 proposals with the top IoU to the ground truth as the positive samples.)
\item $\{{\bf n}^+_1,{\bf n}^+_2,{\bf n}^+_3,{\bf n}^+_4\} = \{21,19,17,15\}$,
abbreviated as $\{21-15\}$
(When the obstacle is in the lowest vertical range,
It selects 21 proposals with the top IoU to the ground truth as the positive samples, and 17 corresponds to the highest vertical range.)
\item $\{{\bf n}^+_1,{\bf n}^+_2,{\bf n}^+_3,{\bf n}^+_4\} = \{15,17,19,21\}$,
abbreviated as $\{15-21\}$
(following the same methods as the variants above.)
\end{itemize}
Based on the results of this experiment,
the positive sample ratio of the best variant is utilized in the later experiment.

Finally,
the ORDR and OBDR are employed to discover the obstacle.
To simplify the expression,
OAOCC+OLP@k+MS is denoted as BP.
OAOCC+OLP@k is denoted as OP,
which is used in our basic method \cite{ICRA}.
All the regressors are trained by using the best positive sample ratio in the previous experiment.
Three variants, i.e., BP+ORDR@4, OP+ORDR@4, BP+Fusion@4,
are defined to validate the effectiveness of the regression model.
Note that ``Fusion'' denotes the model fusing ORDR and OBDR.

\begin{table}[]
\begin{center}
\caption{
Comparison on average recall of our method with others,
the bold numbers denote the best results
}
\begin{tabular}{@{}lccccc@{}}
\specialrule{0em}{2pt}{2pt}
\toprule
\multirow{2}{*}{Proposal Method} & \multicolumn{5}{c}{Proposal Number}   \\
\specialrule{0em}{2pt}{2pt}
& 200   & 400   & 600   & 800   & 1,000  \\
\specialrule{0em}{1pt}{1pt} 
\midrule
\specialrule{0em}{1pt}{1pt}
Objectness \cite{WO} & 0.004 & 0.005 & 0.005 & 0.005 & 0.005 \\
\specialrule{0em}{1pt}{1pt}
SCG \cite{MCG}            & 0.029 & 0.044 & 0.062 & 0.073 & 0.085 \\
\specialrule{0em}{1pt}{1pt}
Selective Search \cite{SS}   & 0.096 & 0.156 & 0.189 & 0.209 & 0.224 \\
\specialrule{0em}{1pt}{1pt}
Geodesic \cite{Geodesic} & 0.130 & 0.146 & 0.150 & 0.154 & 0.156 \\
\specialrule{0em}{1pt}{1pt}
CIOP \cite{CIOP} & 0.072 & 0.079 & 0.084 & 0.086 & 0.088 \\
\specialrule{0em}{1pt}{1pt}
Edge boxes \cite{EB}   & 0.096 & 0.114 & 0.153 & 0.162 & 0.174 \\
\specialrule{0em}{1pt}{1pt}
Object-level Proposal \cite{OLP} & 0.170 & 0.203 & 0.226 & 0.243 & 0.257 \\
\specialrule{0em}{1pt}{1pt}
OAOCC+OLP@1             & 0.174 & 0.208 & 0.234 & 0.249 & 0.264 \\
\specialrule{0em}{1pt}{1pt}
OAOCC+OLP@4             & 0.211 & 0.247 & 0.268 & 0.287 & 0.300 \\
\specialrule{0em}{1pt}{1pt}
OAOCC+OLP@4+MS          & \textbf{0.229} & \textbf{0.267} & \textbf{0.290} & \textbf{0.308} & \textbf{0.324} \\
\specialrule{0em}{1pt}{1pt} \bottomrule
\end{tabular}
\label{table:PropComp}
\end{center}
\end{table}

\begin{figure*}
\centering
\includegraphics[width=1\linewidth]{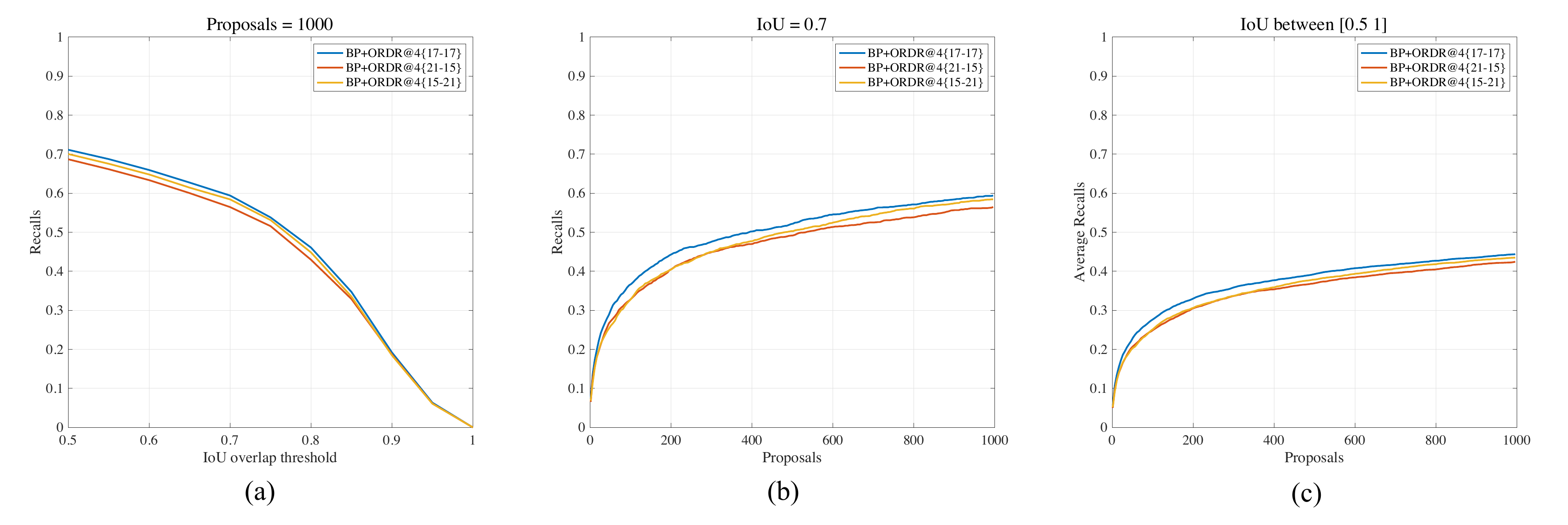}
\caption{
Instance-level recall performance of our variants with different positive sample strategies on the LAFD.
(a) Recall versus the number of proposals given the threshold of IoU is 0.7.
(b) Recall versus IoU threshold given 1,000 proposals.
(c) Average recall versus the number of proposals between IoU 0.5 to 1.}
\label{fig:recal2l}
\end{figure*}

In addition,
for proposal extraction,
$\tau_a$ is set to 6.
$K$ is set to 4.
$\alpha$ is set to 0.65.
For training the ORDR,
the negative sampling number of ORDR $\{{\bf n}^-_1,{\bf n}^-_2,{\bf n}^-_3,{\bf n}^-_4\}$ is set to $\{17,17,17,17\}$.
For OBDR, the number is set to $\{25,25,25,25\}$.
The number of trees in two regressors, namely, $T^{or}$ and $T^{ob}$, is set to 200.
$\tau_l$ is set to 0.2.
$\tau_o$ is set to 0.3.
$\beta_1$ is set to 0.3.
$\tau_\beta$ is set to 0.5.

\subsubsection{Contribution of OAOCC and MS to proposal}
Fig. \ref{fig:recall1} depicts the contribution of OAOCC and MS to the proposal in obstacle discovery.
\textit{edge boxes} (\textit{EB}) and \textit{object-level proposals} (\textit{OLP}) are evaluated with the same image region as OAOCC+OLP@1.
One can see that \textit{OLP} captures more obstacle proposals than \textit{EB},
and OAOCC+OLP@1 outperforms the two methods,
which indicates that the better obstacle contour captured by the obstacle-aware occlusion edge improves the recall of the obstacle proposals.

Then, as the number of ML regions for proposal extraction increases from 1 to 4,
the recall rate of the obstacle proposal also increases;
however, the extent of improvement gradually declines because each ML region contributes to the proposal extraction.

Moreover,
it is noteworthy that OAOCC+OLP@4+MS achieves noticeable improvement over OAOCC+OLP@4,
which demonstrates that the multistride sliding window strategy increases the recall of the obstacle proposals.
The fundamental reason is that hierarchical sampling density corresponds to the obstacle size in each range.
Through the set of experiments,
we also find that the scene prior,
namely, the spatial distribution of the obstacles,
can be used in each module of the algorithm,
and help to improve obstacle discovery.

\begin{table}[!tp]
\begin{center}
\caption{
Comparison on pixel-level ROC of ORDR with different positive sample ratios. The bold numbers denote the best results
}
\begin{tabular}{@{}lcccccc@{}}
\specialrule{0em}{2pt}{2pt}
\toprule
\multirow{2}{*}{\begin{tabular}[c]{@{}l@{}}Positive Samples\\ Ratio\end{tabular}} & \multicolumn{6}{c}{Pixel-level False Positive Rate} \\
\specialrule{0em}{2pt}{2pt}
& 0.005  & 0.010  & 0.015  & 0.020  & 0.025  & 0.030  \\ 
\specialrule{0em}{1pt}{1pt}
\midrule
\specialrule{0em}{1pt}{1pt}
BP+ORDR\{21-15\} & 0.647  & 0.752  & 0.805  & 0.833  & 0.852  & 0.872  \\
\specialrule{0em}{1pt}{1pt}
BP+ORDR\{15-21\} & 0.668  & 0.758  & 0.815  & 0.853  & 0.872  & 0.885  \\
\specialrule{0em}{1pt}{1pt}
BP+ORDR\{17-17\} & \textbf{0.685} & \textbf{0.783} & \textbf{0.833} & \textbf{0.860} & \textbf{0.874} & \textbf{0.892}  \\ 
\specialrule{0em}{1pt}{1pt}
\bottomrule
\end{tabular}
\label{table:roc_2}
\end{center}
\end{table}

\subsubsection{Comparison with other proposal methods}
More methods for extracting object proposals are employed to compare with our method,
i.e., \textit{SCG} (the single scale variant of {\it multiscale combinatorial grouping} \cite{MCG}) and \textit{selective search} \cite{SS}.
The average recall shown in Table \ref{table:PropComp} shows that
our method outperforms other proposal methods by a large margin,
which means that our method exploits the scene prior information to enhance the robustness of the proposal in obstacle discovery.
In conclusion,
for obstacle discovery,
the use of prior scene information must be emphasized.
	
\subsubsection{Choices of positive sample strategy}
In the last experiment,
OAOCC+OLP@4+MS provides the best proposal performance,
which is employed to extract the proposal in the remaining experiments,
abbreviated as BP.
In this section,
we evaluate the impact of the positive sample strategy $\{{\bf n}^+_1,\dots,{\bf n}^+_K\}$ on the regression performance.

\begin{figure}
\centering
\includegraphics[width=1\linewidth]{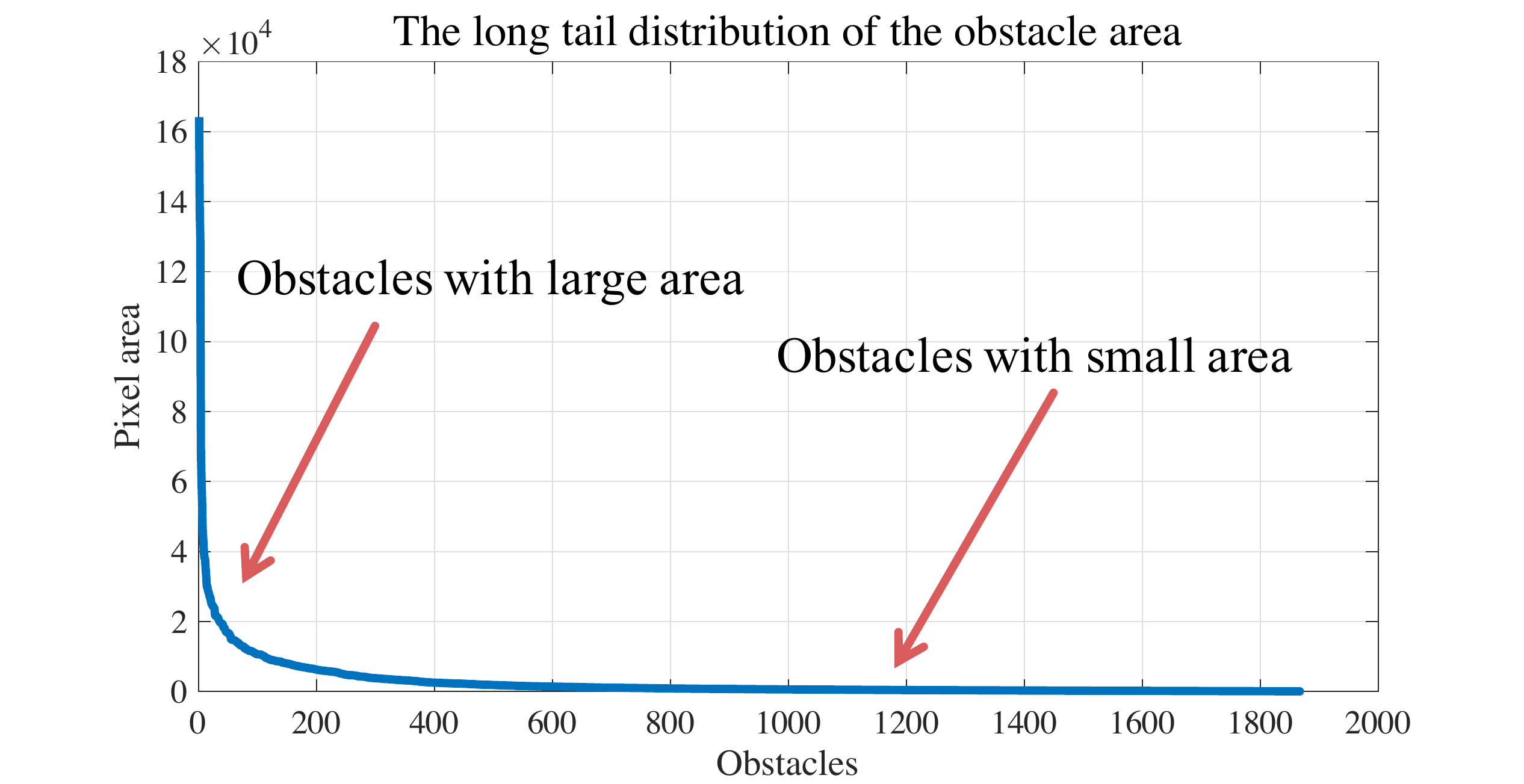}
\caption{
The distribution of the obstacle area in the \textit{Lost and Found} dataset, which follows the long-tail phenomenon.
}
\label{fig:long}
\end{figure}

\begin{figure*}
\centering
\includegraphics[width=1\linewidth]{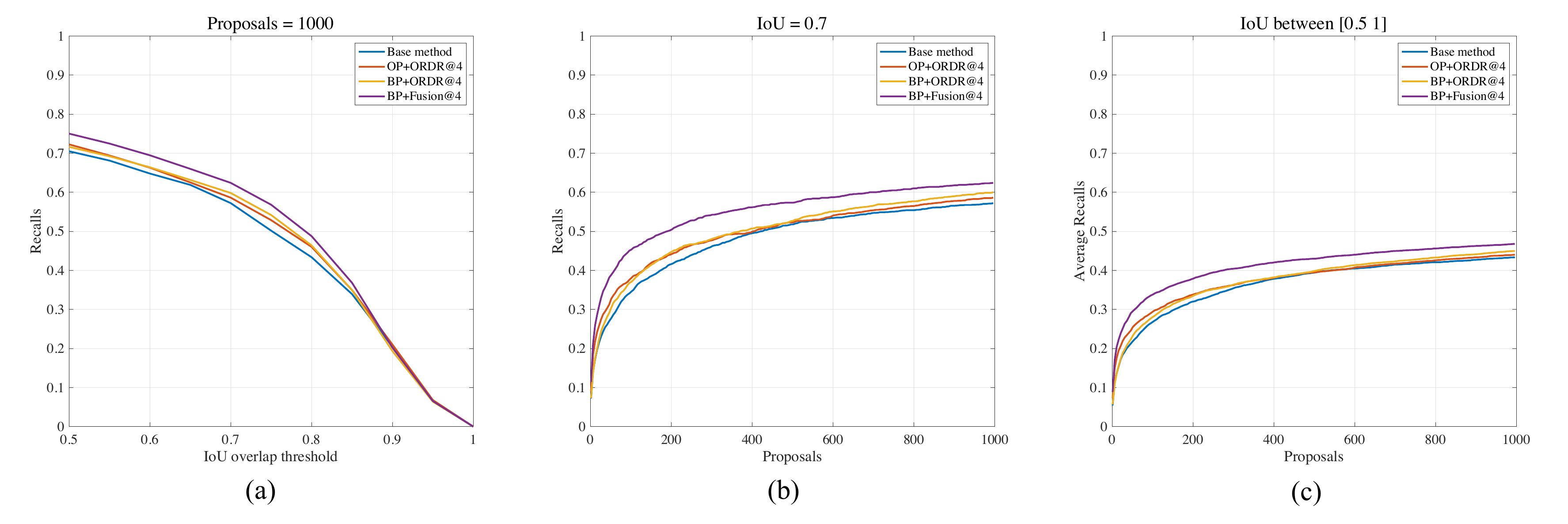}
\caption{
Instance-level recall performance of our variants and the baseline on the LAFD.
(a) Recall versus number of proposals given the threshold of IoU is 0.7.
(b) Recall versus IoU threshold given 1,000 proposals.
(c) Average recall versus number of proposals between IoU 0.5 to 1.}
\label{fig:recall3}
\end{figure*}

Table \ref{table:roc_2} and Fig. \ref{fig:recal2l} depict the results of pixel-level ROC and instance recall, respectively.
BP+ORDR\{21-15\} is prone to learn more obstacles in the short range,
which achieves the worst result.
BP+ORDR\{15-21\} is prone to learn more obstacles
in the long range,
which is better than BP+ORDR\{21-15\}.
In the short range,
the dissimilarity of
the obstacle
with others is sufficiently apparent.
Thus, ORDR easily captures these obstacles in the short range.
However,
one can see that BP+ORDR\{17-17\},
which equally samples the positive proposals in each range,
obtains the best result.
The reason is related to the long-tail distribution in obstacle discovery.
As shown
in Fig. \ref{fig:long},
referring to \cite{SNIP} and \cite{TRID},
the number of small obstacles is much larger than that of the large obstacles.
Thus, BP+ORDR\{15-21\} essentially aggravates the imbalance of samples of different sizes,
which leads to the decision tree selecting too many small obstacle positive samples.
In conclusion, BP+ORDR\{17-17\} is closer to the optimal ratio of sample selection.
As the best variant,
the same sampling parameter is employed in the next comparisons.

\subsubsection{Evaluation of the enhancement to our base algorithm}
To evaluate the enhancement to our base algorithm \cite{ICRA},
Fig. \ref{fig:recall3} and Table \ref{table:roc_3} depict the results of three variants,
i.e., OP+ORDR@4, BP+ORDR@4, and BP+Fusion@4.
It is intuitive that the proposed method outperforms our base algorithm \cite{ICRA}.
In detail,
since more hard negative samples, e.g., scrawl, shadow, and well lid,
are sampled to improve the robustness of ORDR,
OP+ORDR@4 achieves a considerable improvement over our basic method.
BP+Fusion@4 has a visible improvement over BP+ORDR@4 in two metrics.
The improvement intrinsically stems from  OBDR focusing on the dissimilarity between obstacles and other non-obstacles.
It is noteworthy that BP+Fusion@4 achieves 5\% to 10\% accuracy improvement over our basic method,
which is the result of the combination of novel sample selection and OBDR.
However, although BP+ORDR@4 captures the tiny obstacle proposals better,
it is only slightly higher than OP+ORDR@4 in the instance-level recall.
Due to the handcrafted features and learners,
our approach limitedly learns extremely tiny obstacles.

\begin{table}[!tp]
\begin{center}
\caption{Pixel-level ROC of our method and other works on the LAFD, the bold numbers denote the best results}
\begin{tabular}{@{}lcccccc@{}}
\toprule
\specialrule{0em}{2pt}{2pt}
\multirow{2}{*}{Discovery Method} & \multicolumn{6}{c}{Pixel-level False Positive Rate} \\
\specialrule{0em}{2pt}{2pt}
& 0.005  & 0.010  & 0.015  & 0.020  & 0.025  & 0.0319  \\
\specialrule{0em}{1pt}{1pt}
\midrule
\specialrule{0em}{1pt}{1pt}
FPHT-CStix\cite{Pinggera2016Lost}& 0.61 & 0.66 & 0.68 & 0.69 & - & - \\
\specialrule{0em}{1pt}{1pt}
PHT-CStix\cite{Pinggera2016Lost}& 0.62 & 0.66 & 0.67 & 0.68 & - & - \\
\specialrule{0em}{1pt}{1pt}
MergeNet-135\cite{MergeNet}& - & - & - & 0.85 & - & - \\\specialrule{0em}{1pt}{1pt}
MergeNet-1036\cite{MergeNet}& - & - & - & - & - & \textbf{0.929} \\
\specialrule{0em}{1pt}{1pt}
Basic method\cite{ICRA}& 0.620 & 0.752 & 0.798 & 0.850 & 0.857 & - \\
\specialrule{0em}{1pt}{1pt}
OP+ORDR@4               & 0.653 & 0.770 & 0.828 & 0.852 & 0.874 & 0.904 \\
\specialrule{0em}{1pt}{1pt}
BP+ORDR@4               & 0.685 & 0.783 & 0.833 & 0.860 & 0.874 & 0.902 \\
\specialrule{0em}{1pt}{1pt}
BP+Fusion@4        & \textbf{0.722} & \textbf{0.811} & \textbf{0.848} & \textbf{0.873} & \textbf{0.890} & 0.908  \\
\specialrule{0em}{1pt}{1pt}
\bottomrule
\end{tabular}
\label{table:roc_3}
\end{center}
\end{table}
	
\subsubsection{Comparison with other obstacle discovery approaches}
Table \ref{table:roc_3} indicates the comparison of our method against other obstacle discovery methods.
\textit{PHT-CStix} and \textit{FPHT-CStix} \cite{Pinggera2016Lost} discovered the obstacles from the disparity map.
When FPR is equal to 2\%,
our fusion regressor achieves 19\% and 18\% accuracy improvement over \textit{PHT-CStix} and \textit{FPHT-CStix}, respectively.
Similarly,
when FPR is lower,
our method achieves considerable improvement in accuracy over these two methods.
MergeNet-135 \cite{MergeNet} is trained by 135 images and achieves pixel-level accuracy of 85\% with FPR 2.0\%.
MergeNet-1036 achieves a TPR of 92.85\% with a higher FPR.
Although our method has nothing to do with deep learning,
our fusion regressor achieves an approximate result.
	
\begin{figure*}
	\centering
	\includegraphics[width=0.9\linewidth]{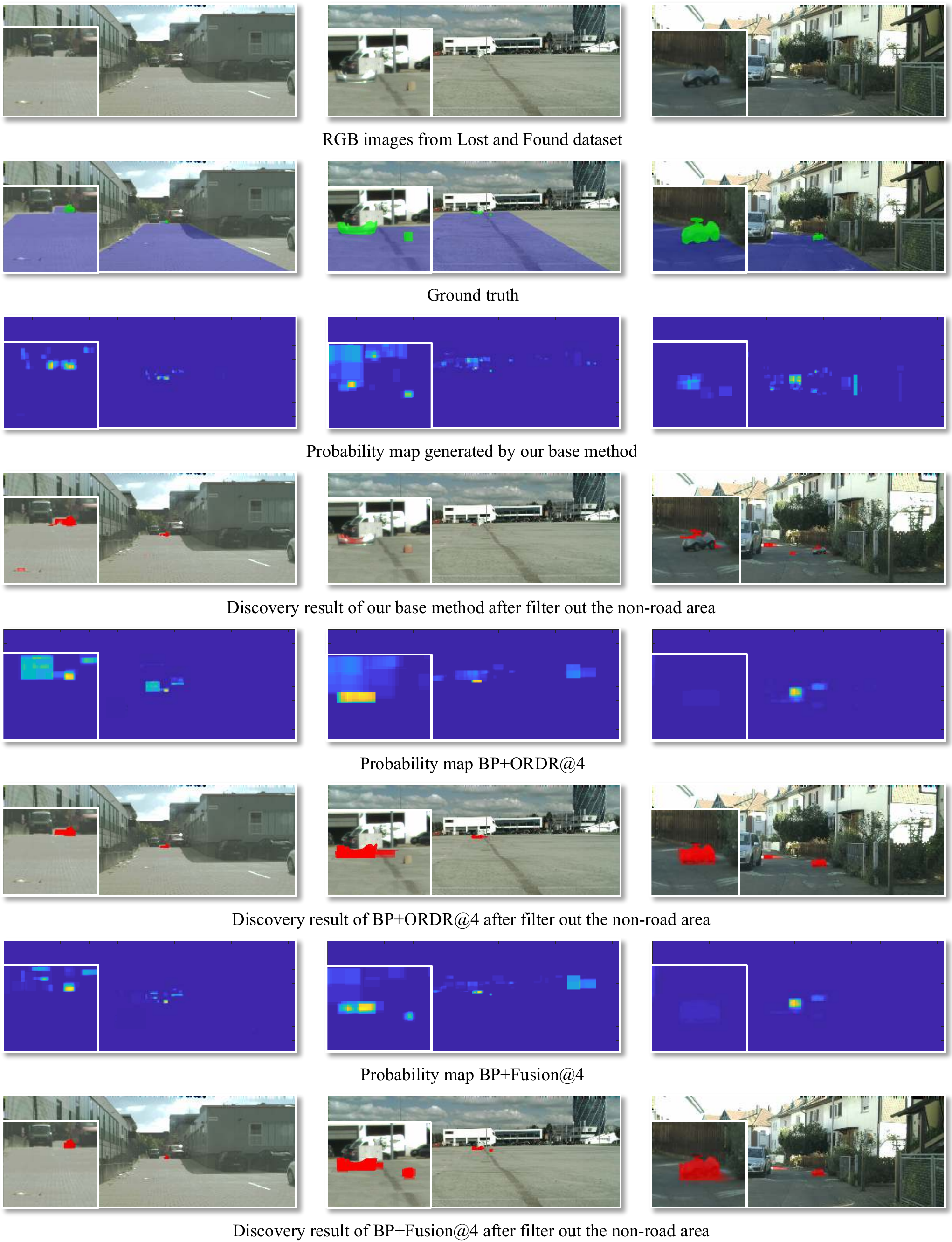}
	\caption{
		Qualitative results for obstacle discovery on several RGB images using our method.
		In the obstacle-occupied probability maps,
		the yellow color corresponds to the high obstacle probability,
		and the blue color corresponds to the low probability.
		In the result,
		the obstacles are marked in red.
		The images can be zoomed to clearly observe these tiny obstacles.
	}
	\label{fig:qualification}
\end{figure*}

\begin{figure*}
		\centering
		\includegraphics[width=1\linewidth]{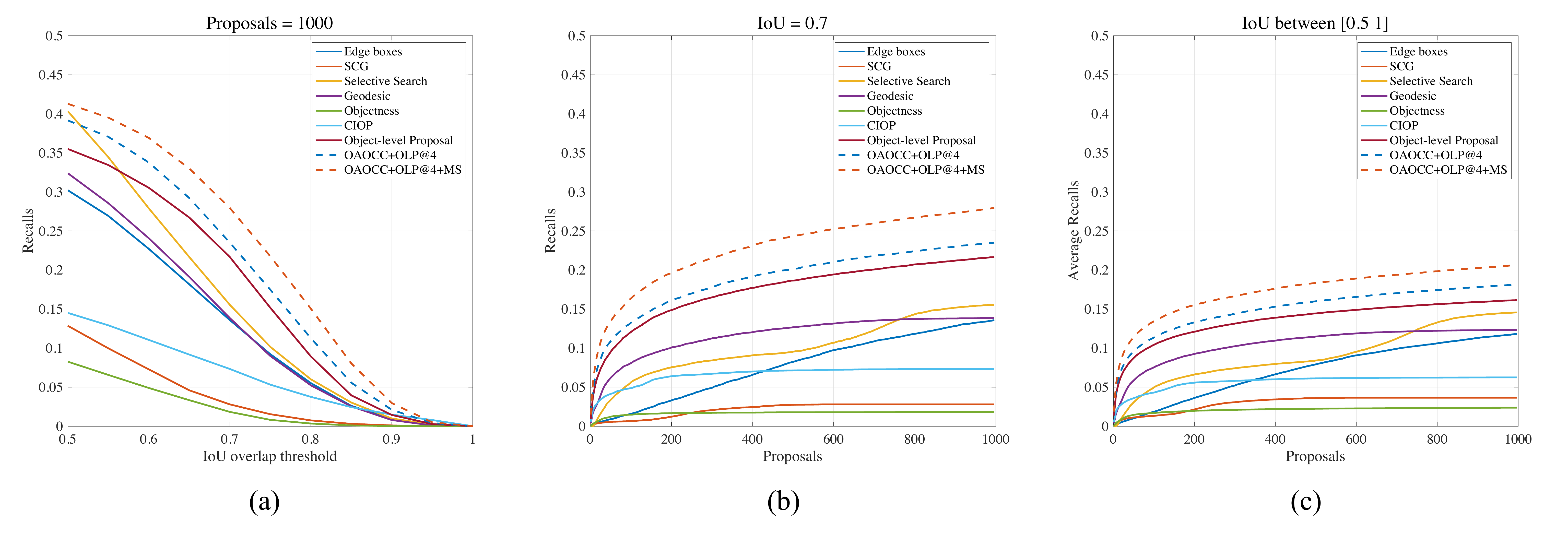}
		\caption{
		Instance-level recall performance of our variants and other methods on MODDv2.
		(a) Recall versus the number of proposals given the threshold of IoU is 0.7.
		(b) Recall versus the IoU threshold given 1,000 proposals.
		(c) Average recall versus the number of proposals between IoU 0.5 and 1.
		}
		\label{fig:recall_modd}
\end{figure*}
	
\subsubsection{Qualification test}
Fig. \ref{fig:qualification} depicts the qualitative results of our method on three challenging scenarios from the testing set.
The left column shows a bobby car with a mixed background in an overexposed scene.
The middle column shows a bumper with a complex background and a box with a road-like color.
The rightmost column shows a bobby car in shadow.
	
In the left column,
the overexposed scene is rare in the training set,
examining the robustness of the proposed
algorithm.
The early version of our method fails to split the obstacle and other objects and generates a small false positive discovery on the road.
In contrast,
the proposed BP+ORDR@4 obtains a higher response for the obstacle and avoids false discovery of the road patch.
Furthermore,
Fusion@4 performs better.
It yields the full discovery of the obstacle,
and avoids false discovery as well.
One can see that there are more proposals aggregated around the obstacle.
The reason is that our secondary regressor learns the dissimilarity between obstacles and non-obstacles,
and thus, gives a higher score to obstacle proposals.
The middle column is also very challenging:
the color of the irregular bumper resembles the car in the background,
and the yellow box resembles a part of the road surface.
Our basic method generates many proposals in the background but fails to discover the bumper
and generates some false positive proposals on the ground.
The proposed BP+ORDR@4 ultimately discovers the irregular bumper but fails to discover the case,
because BP+ORDR@4 considers it the texture of the road.
Fortunately, BP+Fusion@4 successfully discovers the two obstacles.
Intuitively, the proposals of the background are reduced.
In the rightmost column,
the bobby car has a similar HSV color to the road.
Thus, our basic method \cite{ICRA} fails to generate enough obstacle proposals,
and thus loses the obstacle.
Although BP+ORDR@4 obtains a low response on the obstacle,
a full shape is retained,
which helps to discover the obstacle.
Furthermore, BP+Fusion@4 generates more obstacle proposals and thus obtains a relatively higher response in the obstacle area.
Our method completely discovers distant obstacles.

\begin{table}[]
\begin{center}
\caption{
Average recall of our variants and other methods on the MODDv2, the bold numbers denote the best results}
\begin{tabular}{@{}lccccc@{}}
\specialrule{0em}{2pt}{2pt}
\toprule
\multirow{2}{*}{Method} & \multicolumn{5}{c}{Proposal Number}   \\
\specialrule{0em}{2pt}{2pt}
& 200   & 400   & 600   & 800   & 1,000  \\
\specialrule{0em}{1pt}{1pt} 
\midrule
\specialrule{0em}{1pt}{1pt}
Objectness \cite{WO} & 0.020 & 0.022 & 0.023 & 0.023 & 0.024 \\
\specialrule{0em}{1pt}{1pt}
SCG \cite{MCG}            & 0.021 & 0.034 & 0.036 & 0.036 & 0.037 \\
\specialrule{0em}{1pt}{1pt}
Selective Search \cite{SS}   & 0.066 & 0.079 & 0.095 & 0.133 & 0.146 \\
\specialrule{0em}{1pt}{1pt}
Geodesic \cite{WO}  & 0.093 & 0.109 & 0.119 & 0.122 & 0.123 \\
\specialrule{0em}{1pt}{1pt}
CIOP \cite{CIOP}  & 0.056 & 0.060 & 0.062 & 0.062 & 0.062 \\
\specialrule{0em}{1pt}{1pt}
Edge boxes \cite{EB}   & 0.036 & 0.066 & 0.091 & 0.106 & 0.118 \\
\specialrule{0em}{1pt}{1pt}
Object-level Proposal \cite{OLP} & 0.121 & 0.139 & 0.149 & 0.156 & 0.161 \\
\specialrule{0em}{1pt}{1pt}
OAOCC+OLP@4             & 0.133 & 0.153 & 0.166 & 0.174 & 0.181 \\
\specialrule{0em}{1pt}{1pt}
OAOCC+OLP@4+MS          & \textbf{0.155} & \textbf{0.176} & \textbf{0.189} & \textbf{0.198} & \textbf{0.206} \\
\specialrule{0em}{1pt}{1pt}
\midrule
\specialrule{0em}{1pt}{1pt}
SSM \cite{cyber_FIOD} & \multicolumn{5}{c}{recall=0.0089 when IoU=0.5} \\
\specialrule{0em}{1pt}{1pt}
\midrule
\specialrule{0em}{1pt}{1pt}
Basic method \cite{ICRA} & 0.151 & 0.175 & 0.185 & 0.191 & 0.195 \\
\specialrule{0em}{1pt}{1pt}
BP+ORDR@4          & 0.200 & 0.231 & 0.247 & 0.258 & 0.267 \\
\specialrule{0em}{1pt}{1pt}
BP+Fusion@4        & \textbf{0.211} & \textbf{0.243} & \textbf{0.259} & \textbf{0.270} & \textbf{0.278} \\
\specialrule{0em}{1pt}{1pt} \bottomrule
\end{tabular}
\label{table:recall_modd}
\end{center}
\end{table}

\subsection{Marine Obstacle Detection Dataset}
\subsubsection{Dataset}
Different from the LAFD,
the Marine Obstacle Detection Dataset 2 (MODDv2) \cite{BOVCON2018} is a realistic,
publicly available Unmanned Surface Vehicle (USV) dataset,
which is also the largest dataset in this scenario.
It comprises IMU data and annotated stereo videos in which a polygon annotates the water,
and a bounding box annotates an obstacle.
In addition,
the whole dataset contains 28 video sequences with different lengths,
a total of 11,675 frames with a resolution of 1,278 $\times$ 958 pixels.
To fully evaluate our method with more test data,
the dataset is randomly divided into 70\% sequences for testing and 30\% sequences for training.
Moreover, only the left camera is utilized to verify our method.

\begin{figure*}
		\centering
		\includegraphics[width=1\linewidth]{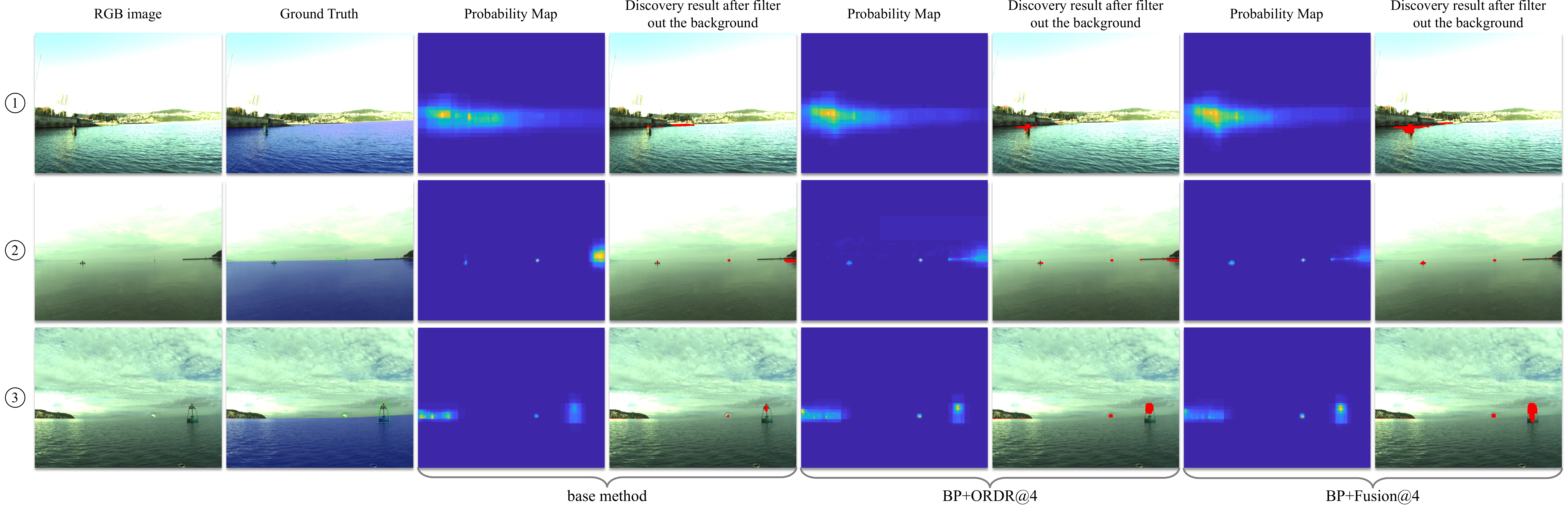}
		\caption{Qualitative results for obstacle discovery on the MODD v2.
		The obstacle in the ground truth is marked in green, and the prediction is marked in red.
		}
		\label{fig:img_modd}
\end{figure*}

\subsubsection{Variants}
In the last experiment,
the proposed variants OAOCC+OLP@4+MS, OAOCC+OLP@4, BP+ORDR@4, and BP+Fusion@4 achieve the best results in various stages.
Thus, their performances are verified on MODDv2.

\begin{table}[!tp]
\begin{center}
\caption{Pixel-level ROC of our variants on the MODDv2, the bold numbers denote the best results}
\begin{tabular}{@{}lcccccc@{}}
\toprule
\specialrule{0em}{2pt}{2pt}
\multirow{2}{*}{Method} & \multicolumn{6}{c}{Pixel-level False Positive Rate} \\
\specialrule{0em}{2pt}{2pt}
& 0.005  & 0.010  & 0.015  & 0.020  & 0.025  & 0.030  \\
\specialrule{0em}{1pt}{1pt}
\midrule
\specialrule{0em}{1pt}{1pt}
Basic method \cite{ICRA}& 0.413  & 0.549  & 0.627  & 0.678  & 0.714  & 0.742      \\
\specialrule{0em}{1pt}{1pt}
BP+ORDR@4               & \textbf{0.464}  & \textbf{0.595}  & 0.666  & 0.714  & 0.748  & 0.773  \\
\specialrule{0em}{1pt}{1pt}
BP+Fusion@4        & 0.449 & 0.591 & \textbf{0.677} & \textbf{0.731} & \textbf{0.768} & \textbf{0.795}  \\
\specialrule{0em}{1pt}{1pt}
\bottomrule
\end{tabular}
\label{table:roc_modd}
\end{center}
\end{table}

\subsubsection{Quantitative results}
The pixel-level quantitative results of these variants are shown in Table \ref{table:roc_modd}.
Although the performance gap between the three variants is small,
BP+Fusion@4 achieves the best performance.
When FPR is 0.03,
the BP+ORDR@4 achieves 3.1\% improvement in pixel-level accuracy,
and Fusion+ORDR@4 corresponds to 5.3\%.
Since our algorithm cannot solve the misdetection to reflection,
the scenario with reflection easily leads to an unfair comparison between variants;
thus, it is removed in testing.

The instance-level results are shown in Fig. \ref{fig:recall_modd} and Table \ref{table:recall_modd}.
The semantic segmentation method (SSM) \cite{cyber_FIOD} only achieves 0.0089 of recall with the default parameter when the IoU threshold is 0.5.
Intuitively,
OAOCC+OLP@4 and OAOCC+OLP@4+MS achieve the best performance.
With pixelwise segmentation of the image,
the grouping-based methods (e.g., \textit{SCG} \cite{MCG}, \textit{selective search} \cite{SS}, and \textit{geodesic} \cite{Geodesic}) usually find the object effectively.
However,
due to the tiny size of the obstacle (such as 7 $\times$ 8 in an image with 1,278 $\times$ 958 pixels),
a sufficiently accurate segmentation cannot be obtained by acceptable computation.
Thus, their performance is limited, as shown in Table \ref{table:recall_modd}.
Based on the window scoring methods (namely, \textit{edge boxes} and \textit{OLP}),
OAOCC+OLP@4 improves the edge probability of tiny obstacles,
hence discovering more obstacles than the two-based methods.
OAOCC+OLP@4+MS performs pixel-by-pixel sampling in areas with tiny obstacles
and is more than 2\% higher than OAOCC+OLP@4 in the average recall.

Based on the multistride sampling strategy and more reasonable training sample selection,
BP+ORDR@4 is slightly ahead of the basic method.
By further distinguishing background and obstacles,
BP+Fusion@4 discovers more obstacles than ORDR alone.

\begin{table}[!tp]
\begin{center}
\caption{Time analysis in processing a single image, - denotes empty, the bold numbers indicate the main computational costs}
\begin{tabular}{@{}|c|c|c|c|@{}}
\hline
Module & Time & Basic method & Time of basic method\\
\hline
OAOCC &\textbf{2.08s}& Occlusion \cite{IS}&\textbf{2.07s}\\
\hline
MS-OLP & \textbf{8.10s} & OLP \cite{OLP} & \textbf{8.10s} \\
\hline
Feature &0.68s&-&-\\
\hline
Regressor &1.14s&-&-\\
\hline
Probability map &0.03s&-&-\\
\hline
All &12.03s&-&10.17s\\
\hline
\end{tabular}
\label{table:modtime}
\end{center}
\end{table}

\subsubsection{Qualitative test}
Fig. \ref{fig:img_modd} illustrates the qualitative results of all methods in the marine scene.
The first row depicts an obstacle when the land is a complex background.
The second row depicts several tiny obstacles.
The third column depicts a hollow buoy.
In the first scene,
the land is easily considered as an obstacle,
making the real obstacle proposal difficult to find.
Due to the lack of distinguishing between obstacles and backgrounds,
our baseline and BP+ORDR@4 fails to fully segment the obstacle.
One can see that the proposed BP+Fusion@4 alleviates this issue.
In the second scene,
our basic method \cite{ICRA} predicts the land as an obstacle.
BP+ORDR@4 alleviates misdetection but fails to fully cover the tiny obstacle.
BP+Fusion@4 discovers the two obstacles,
illustrating the effectiveness of our method in tiny obstacle discovery.
In the third column,
due to the hollow buoy,
it is difficult to fully segment the obstacle.
Although all methods discover the obstacles,
our basic method lost the body of the buoy,
which might lead to collisions in driving.
The BP+ORDR@4 achieves better performance than the basic method,
However, BP+Fusion@4 covers obstacles almost completely.

\begin{table}[!tp]
\begin{center}
\caption{Comparison on time per image, - denotes unknown}
\vspace{-8pt}
\begin{tabular}{@{}|l|c|c|c|c|@{}}
\hline
 & PHT&FPHT&MergeNet&Ours \\
\hline
Time per image &0.5 s&0.05 s&0.2 s&12.03 s\\
\hline
Hardware &GPU(-)&GPU(-)&GPU(1080Ti)&CPU(i7-2600k)\\
\hline
Implementation &-&-&-&MATLAB\\
\hline
Resolution &full&full&-&full\\
\hline
\end{tabular}
\label{table:time}
\end{center}
\end{table}

\subsection{Computational performance analysis}
Computational performance is also an essential factor in evaluating the tiny obstacle discovery.
For our method,
the computational cost mainly consists of five parts:
obstacle-aware occlusion edge (OAOCC),
multistride object proposal extraction (MS-OLP),
feature vector calculation,
proposal regression,
and obstacle-occupied probability map generation.
The time per part is shown in Table \ref{table:modtime}.
In detail,
the first part is based on \textit{occlusion edge} \cite{IS},
and extra utilizes several matrix additions.
Thus, it has the same time complexity as \cite{IS}.
Numerically, it costs $2.08$s,
which is the same as \cite{IS}.
Secondly,
based on \textit{OLP} \cite{OLP} whose time complexity is $\mathcal{O}(n)$,
the time complexity of MS-OLP with four ML regions is also $\mathcal{O}(n)$,
where $n$ is the number of proposals in each region.
MS-OLP costs $8.10$ s,
which is the same as \cite{OLP} and occupies the main computational cost.
For feature vector calculation,
the time complexity of the feature using an integral map is $m^1\mathcal{O}(n)$,
and that of the feature without an integral map is $m^2\mathcal{O}(nl)$,
where $m^1$ is the number of features using an integral map,
$m^2$ corresponds to the features without using an integral map,
$n$ is the number of proposals, and
$l$ is the number of pixels inside the proposal.
This part takes $0.68$ s.
For proposal regression,
the time complexity is $\mathcal{O}(Thn)$,
where $T=T^{or}+T^{ob}$ is the number of trees,
and $h$ is the depth of each tree.
This part takes $1.14$ s.
Finally,
for obstacle-occupied probability map generation,
the time complexity is $\mathcal{O}(nl)$,
and the time cost is $0.03$ s.
In summary,
Although the first two parts take up most of the overall time cost, namely, 10.18 s,
it is apparent that their basic algorithms,
namely, occlusion edge \cite{IS} and OLP \cite{OLP},
are the intrinsic factors for time consumption.
In contrast, the remaining modules take much less time than the first two modules,
only costing 1.85 s.
To overcome this challenge,
we will improve the effectiveness of the first two parts in future work.

Table \ref{table:time} depicts the comparison of time per image between our method and the previous method.
Since \cite{MergeNet,Pinggera2016Lost} does not provide complete information about the computing platform,
it is difficult to make a comprehensive comparison.
Our method is implemented in MATLAB and tested on a PC with 16 GB memory and an Intel Core i7-2600K CPU,
and takes more time to process an image.
As far as we know,
there are two main reasons:
\begin{itemize}
\item The advantage of the scene prior is not fully utilized to reduce the computational cost of the first two parts of our method.
\item Lack of parallel computing to optimize the algorithm.
\end{itemize}

\section{Conclusion and Future Work}
In this paper, a novel obstacle discovery method is introduced.
This method applies the scene prior information to construct the multilayer region and utilizes it in each module of the obstacle discovery.
Firstly, the novel obstacle-aware occlusion edge map is produced by enhancing the edge cues in each multilayer region.
Secondly, a multistride sliding window strategy is proposed to ensure the existence of tiny obstacle proposals.
In addition, an obstacle-aware regression model,
which consists of a primary-secondary regressor,
is introduced to generate the obstacle-occupied probability map.
Finally,
extensive experiments validate the effectiveness of each part of the proposed method.
	
In future work,
we will further incorporate the scene priors to constrain the edge detection,
reduce the redundant sliding window while obtaining proposals,
and eventually, improve the effectiveness of the entire method.
Referring to \cite{MergeNet,Pinggera2016Lost},
we will parallelize our algorithm by GPU to handle the heavy computational cost during box scoring and feature extraction.

\ifCLASSOPTIONcaptionsoff
\newpage
\fi

\bibliographystyle{IEEEtran}
\bibliography{TOD_final}

\begin{IEEEbiography}[{\includegraphics[width=1in,height=1.25in,clip,keepaspectratio]{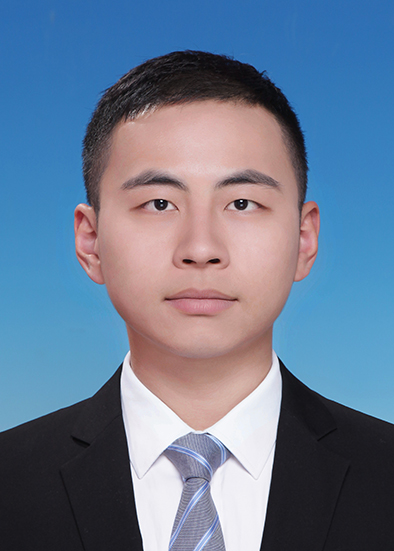}}]{Feng Xue} received the M.S. degree in  Beijing University of Posts and Telecommunications (BUPT), Beijing, China, in 2019.
He is currently a Ph.D candidate majoring in BUPT. His research interests include pattern recognition, image processing and visual obstacle discovery of autonomous vehicles.

\end{IEEEbiography}

\begin{IEEEbiography}[{\includegraphics[width=1in,height=1.25in,clip,keepaspectratio]{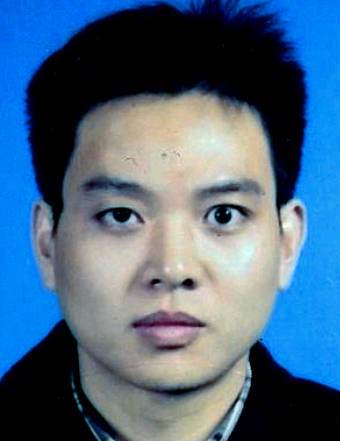}}]{Anlong Ming}
received his B.S. degrees from Wuhan University in 2001,
and Ph.D. degree in Beijing University of Posts and Telecommunications in 2008.
He is currently a professor with the School of Computer Science,
Beijing University of Posts and Telecommunications, Beijing, China.
His research interests include computer vision and robot vision.
E-mail: mal@bupt.edu.cn
\end{IEEEbiography}
	
\begin{IEEEbiography}[{\includegraphics[width=1in,height=1.25in,clip,keepaspectratio]{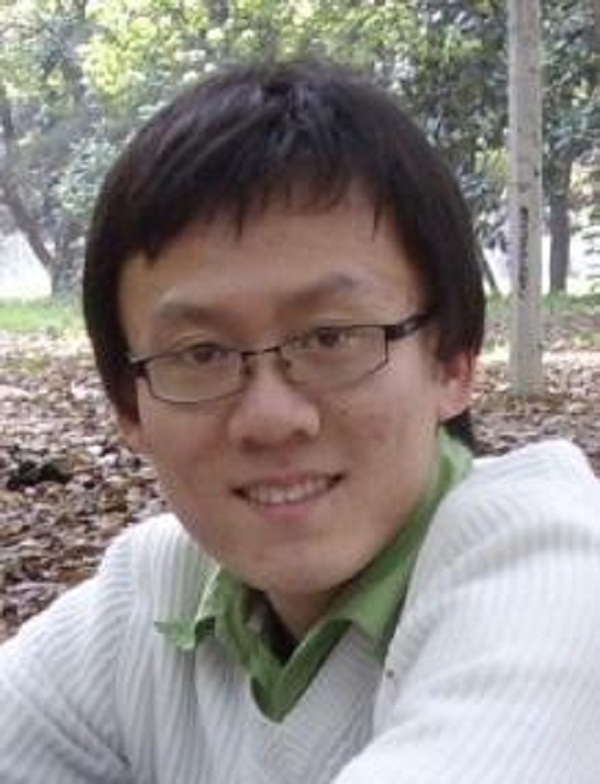}}]{Yu Zhou}
received the M.S. and Ph.D. degrees both in Electronics and Information Engineering from Huazhong University of Science and Technology (HUST), Wuhan, P.R. China in 2010, and 2014, respectively.
In 2014, he joined the Beijing University of Posts and Telecommunications (BUPT), Beijing, as a Postdoctoral Researcher from 2014 to 2016, an Assistant Professor from 2016 to 2018.
He is currently an Associate Professor with the School of Electronic Information and Communications, HUST.
His research interests include computer vision and automatic drive.
\end{IEEEbiography}

\end{document}